\documentclass{article}

\usepackage{microtype}
\usepackage{subfig}
\usepackage{booktabs}
\usepackage{stfloats}
\usepackage{makecell}
\usepackage{enumitem}

\usepackage[accepted]{icml2021} 
\bibpunct{(}{)}{,}{a}{}{;}
\usepackage{graphicx}
\usepackage{amsmath,amssymb,amsfonts}
\usepackage{amsthm}
\usepackage{thmtools}
\usepackage{float}
\usepackage{pifont}
\newcommand{\cmark}{\ding{51}}%
\newcommand{\xmark}{\ding{55}}%
\usepackage{amsthm}
\usepackage{dsfont}
\usepackage{epigraph}

\usepackage{tikz}
\usetikzlibrary{arrows,calc} 
\newcommand{\tikzAngleOfLine}{\tikz@AngleOfLine}
\def\tikz@AngleOfLine(#1)(#2)#3{%
\pgfmathanglebetweenpoints{%
\pgfpointanchor{#1}{center}}{%
\pgfpointanchor{#2}{center}}
\pgfmathsetmacro{#3}{\pgfmathresult}%
}
\newtheorem{definition}{Definition}
\newtheorem{theorem}{Theorem}

\newtheorem{lemma}{Lemma}
\declaretheoremstyle[
headfont=\normalfont\itshape,
qed=\qedsymbol,
]{mypf}
\declaretheorem[numbered=no, name=Proof, style=mypf]{pf}

\usepackage{hyperref}
\hypersetup{bookmarks=true}
\usepackage{color-edits}
\addauthor{sw}{blue}
\addauthor{gs}{violet}

\usepackage{xcolor}
\definecolor{expert}{HTML}{008000}
\definecolor{error}{HTML}{f96565}

\newcommand{\figref}[1]{Fig. \ref{#1}}
\usepackage[font=small,labelfont=bf]{caption}
\def\BibTeX{{\rm B\kern-.05em{\sc i\kern-.025em b}\kern-.08em
    T\kern-.1667em\lower.7ex\hbox{E}\kern-.125emX}}
\DeclareMathOperator*{\argmax}{arg\,max}
\DeclareMathOperator*{\argmin}{arg\,min}
\usepackage[normalem]{ulem}
\newcommand{\overbar}[1]{\mkern 1.5mu\overline{\mkern-1.5mu#1\mkern-1.5mu}\mkern 1.5mu}

\icmltitlerunning{Of Moments and Matching}
\begin{document}

\twocolumn[
\icmltitle{Of Moments and Matching: \\ A Game-Theoretic Framework for Closing the Imitation Gap}




\begin{icmlauthorlist}
\icmlauthor{Gokul Swamy}{ri}
\icmlauthor{Sanjiban Choudhury}{aurora}
\icmlauthor{J. Andrew Bagnell}{ri,aurora}
\icmlauthor{Zhiwei Steven Wu}{isr}
\end{icmlauthorlist}

\icmlaffiliation{ri}{Robotics Institute, Carnegie Mellon University}
\icmlaffiliation{isr}{Institute for Software Research, Carnegie Mellon University}
\icmlaffiliation{aurora}{Aurora Innovation}

\icmlcorrespondingauthor{Gokul Swamy}{gswamy@cmu.edu}

\icmlkeywords{Imitation Learning, Reinforcement Learning, Robotics, GANs}

\vskip 0.3in
]

\printAffiliationsAndNotice{} 



\begin{abstract}
We provide a unifying view of a large family of previous imitation learning algorithms through the lens of \textit{moment matching}. At its core, our classification scheme is based on whether the learner attempts to match (1) \textit{reward} or (2) \textit{action-value} moments of the expert's behavior, with each option leading to differing algorithmic approaches. By considering adversarially chosen divergences between learner and expert behavior, we are able to derive bounds on policy performance that apply for all algorithms in each of these classes, the first to our knowledge. We also introduce the notion of \textit{moment recoverability}, implicit in many previous analyses of imitation learning, which allows us to cleanly delineate how well each algorithmic family is able to mitigate compounding errors. We derive three novel algorithm templates (\texttt{AdVIL}, \texttt{AdRIL}, and \texttt{DAeQuIL}) with strong guarantees, simple implementation, and competitive empirical performance.

\end{abstract}

\section{Introduction}
\label{sec:intro}

\begin{figure}
    \centering
    \includegraphics[width=\columnwidth]{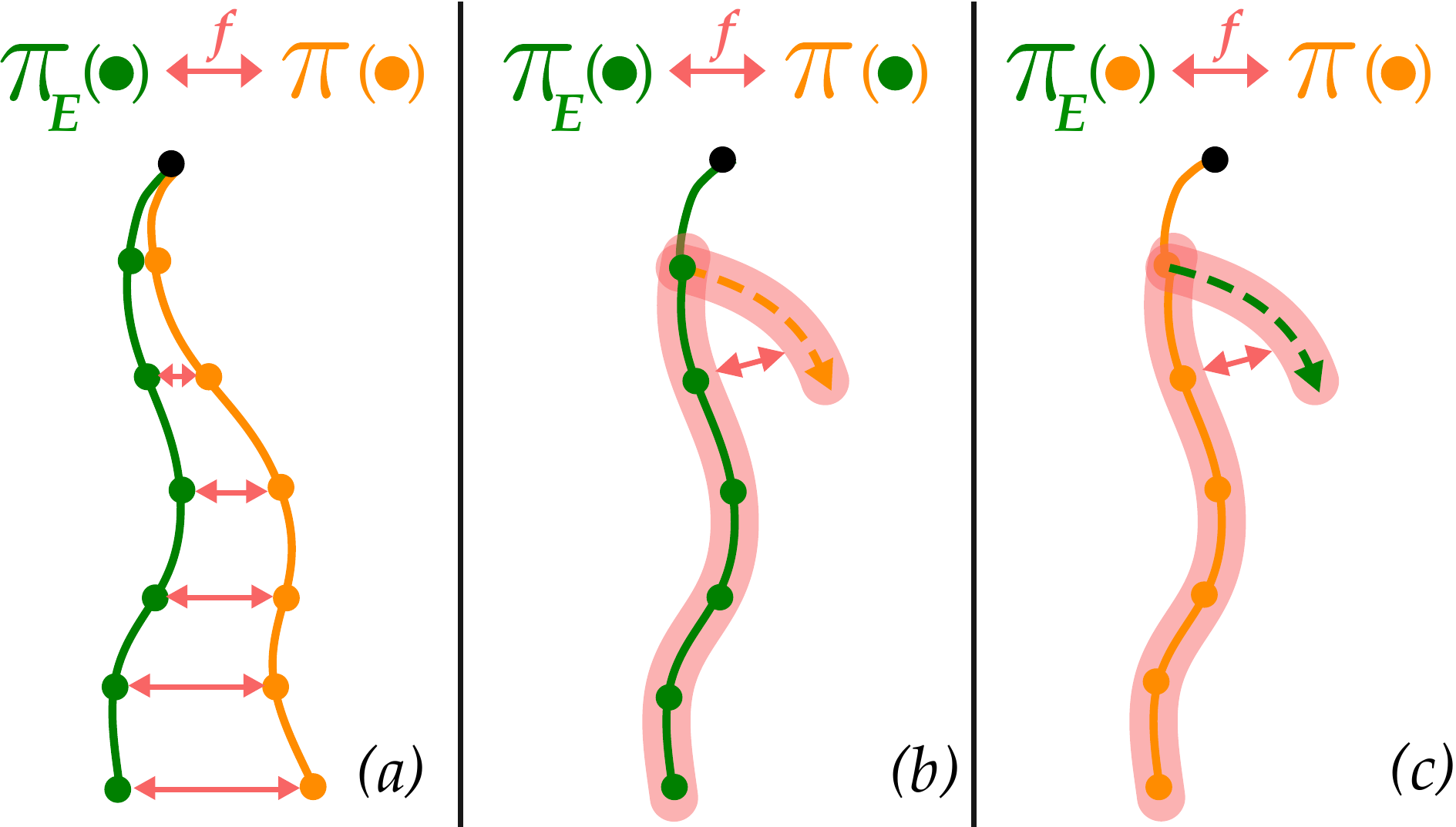}
    \caption{We consider three classes of imitation learning algorithms. \textbf{(a)} On-policy reward moment-matching algorithms require access to the environment to generate learner trajectories. \textbf{(b)} Off-policy $Q$-value moment-matching algorithms run completely offline but can produce policies with quadratically compounding errors. \textbf{(c)} On-policy $Q$-value moment-matching algorithms require access to the environment and a queryable expert but can produce strong policies in \textit{recoverable} MDPs.}

    \label{fig:front}\vspace{-1em}
\end{figure}
When formulated as a statistical learning problem, imitation learning is fundamentally concerned with finding policies that minimize some notion of divergence between learner behavior and that of an expert demonstrator. Existing work has explored various types of divergences including KL \cite{pomerleau1989alvinn, DBLP:journals/corr/BojarskiTDFFGJM16}, Jensen-Shannon \cite{rhinehart2018r2p2}, Reverse KL \cite{kostrikov2019imitation}, $f$ \cite{DBLP:journals/corr/abs-1905-12888}, and Wasserstein \cite{dadashi2020primal}. 

At heart, though, we care about the performance of the learned policy under an objective function that is not known to the learner. As argued by \cite{abbeel2004apprenticeship} and \cite{ziebart2008maximum}, this goal is most cleanly formulated as a problem of \textit{moment matching}, or, equivalently, optimizing Integral Probability Metrics \cite{sun2019provably} (IPMs). This is because a learner that in expectation matches the expert on all the basis functions of a class that includes the expert's objective function, or \textit{matches moments}, must achieve the same performance and will thus be indistinguishable in terms of quality. Additionally, in sharp contrast to recently proposed approaches \cite{DBLP:journals/corr/abs-1905-12888, kostrikov2019imitation, jarrett2020strictly, rhinehart2018r2p2}, moments, due to their simple forms as expectations of basis functions, can be effectively estimated via demonstrator samples and the uncertainty in these estimates can often be quantified to regularize the matching objective \cite{dudik2004performance}. In short: these moment matching procedures are simple, effective, and provide the strongest policy performance guarantees we are aware of for imitation learning. 

As illustrated in \figref{fig:front}, there are three classes of moments a learner can focus on matching: (a) \textit{on-policy reward moments}, (b) \textit{off-policy $Q$-value moments}, and (c) \textit{on-policy $Q$-value moments}, each of which have different requirements on the environment and on the expert. We abbreviate them as \textbf{reward}, \textbf{off-$\mathbf{Q}$}, and \textbf{on-$\mathbf{Q}$} moments, respectively.



Our key insight is that reward moments have more \textit{discriminative power} because they can pick up on differences in induced state visitation distributions rather than just action conditionals. Thus, \textit{reward moment matching is a harder problem with stronger guarantees than off-$Q$ and on-$Q$ moment matching}.

Our work makes the following three contributions:

\noindent \textbf{1. We present a unifying framework for moment matching in imitation learning.} Our framework captures a wide range of prior approaches and allows us to construct, to our knowledge, the first formal lower bounds demonstrating that the choice between matching ``reward'', ``off-$Q$'', or ``on-$Q$'' moments is fundamental to the problem of imitation learning rather than an artifact of a particular algorithm or analysis.

\noindent \textbf{2. We clarify the dependence of imitation learning bounds on problem structure.} We introduce a joint property of an expert policy and moment class, \textit{moment recoverability}, that helps us characterize the problems for which compounding errors are likely to occur, regardless of the kind of feedback the learner is exposed to.

\noindent \textbf{3. We provide three novel algorithms with strong performance guarantees.} We derive idealized algorithms that match each class of moments. We also provide practical instantiations of these ideas, \texttt{AdVIL}, \texttt{AdRIL}, and \texttt{DAeQuIL}. These algorithms have significantly different practical performance as well as theoretical guarantees in terms of compounding of errors over time steps of the problem.

\section{Related Work}
\label{sec:related}

\noindent \textbf{Imitation Learning.}
Imitation learning has been shown to be an effective method of solving a variety of problems, from getting cars to drive themselves \cite{pomerleau1989alvinn}, to achieving superhuman performance in games like Go \cite{silver2016mastering, sun2019dual}, to sample-efficient learning of control policies for high DoF robots \cite{levine2013guided}, to allowing human operators to effectively supervise and teach robot fleets \cite{swamy2020scaled}. The issue of compounding errors in imitation learning was first formalized by \cite{ross2011reduction}, with the authors proving that an interactive expert that can suggest actions in states generated via learner policy rollouts will be able to teach the learner to recover from mistakes. 



\noindent \textbf{Adversarial Imitation Learning.} Starting with the seminal work of  \cite{ho2016generative}, numerous proposed approaches have framed imitation learning as a game between a learner's policy and another network that attempts to discriminate between learner rollouts and expert demonstrations \cite{fu2018learning, DBLP:journals/corr/abs-1807-09936}. We build upon this work by elucidating the properties that result from the kind of \textit{feedback} the learner is exposed to -- whether they are able to see the consequences of their own actions via rollouts or if they are only able to propose actions in states from expert trajectories. Our proposed approaches also have stronger guarantees and less brittle performance than the popular GAIL \cite{ho2016generative}.


\noindent \textbf{Mathematical Tools.} Our algorithmic approach combines two tools that have enjoyed success in imitation learning: \textit{functional gradients} \cite{ratliff2009learning} and the \textit{Integral Probability Metric} \cite{sun2019provably}. We define two algorithms, \texttt{AdRIL} and \texttt{AdVIL} that are based on optimizing the value-directed IPM, with \texttt{AdRIL} having the discriminative player perform updates via functional gradient descent. The IPM is linear in the discriminative function, unlike other proposed metrics like the Donsker-Varadhan bound on KL divergence. Specifically, the Donsker-Varadhan bound includes an expectation of the exponentiated discriminative function, which makes estimation difficult with a few samples \cite{mcallester2020formal}. Our analysis makes repeated use of the \textit{Performance Difference Lemma} \cite{kakade2002approximately, bagnell2003policy} or PDL, which allows us to bound the suboptimality of the learner's policy.


Our proposed algorithms bear some resemblance to previously proposed methods, with \texttt{AdRIL} resembling SQIL \cite{reddy2019sqil} and \texttt{AdVIL} resembling ValueDICE \cite{kostrikov2019imitation}. We note that \texttt{AdVIL}, while cleanly derived from the PDL, can also be derived from an IPM by using a telescoping substitution similar to the ValueDICE derivation. Notably, because \texttt{AdVIL} is linear in the discriminator, it does not suffer from ValueDICE's difficulty in estimating the expectation of an exponential. This difficulty might help explain why ValueDICE can underperform the behavioral cloning baseline on several benchmark tasks \cite{jarrett2020strictly}. Similarly, \texttt{AdRIL} can avoid the sharp degradation in policy performance that SQIL demonstrates \cite{barde2020adversarial}. This is because SQIL hard-codes the discriminator while \texttt{AdRIL} adaptively updates the discriminator to account for changes in the policy's trajectory distribution. \texttt{DAeQuIL} can be seen as the natural extension of DAgger \cite{ross2011reduction} to the adversarial loss setting.

\section{Moment Matching Imitation Learning}
We begin by formalizing our setup and objective.
\label{sec:mmil}
\subsection{Problem Definition}
Let $\Delta(\mathcal{X})$ denote the space of all probability distribution over a set $\mathcal{X}$. Consider an MDP parameterized by $\langle \mathcal{S}, \mathcal{A}, \mathcal{T}, r, T, P_0 \rangle$\footnote{We ignore the discount factor for simplicity.}, where $\mathcal{S}$ is the state space, $\mathcal{A}$ is the action space, $\mathcal{T}: \mathcal{S} \times \mathcal{A} \rightarrow \Delta(\mathcal{S})$ is the transition operator, $r: \mathcal{S} \times \mathcal{A} \rightarrow [-1, 1]$ is a reward function, $T$ is the horizon, and $P_0$ is the initial state distribution.
Let $\pi$ denote the learner's policy and let $\pi_E$ denote the demonstrator's policy. A trajectory $\tau \sim \pi = \{s_t, a_t\}_{t=1 \dots T}$ refers to a sequence of state-action pairs generated by first sampling a state $s_1$ from $P_0$ and then repeatedly sampling actions $a_t$ and next states $s_{t+1}$ from $\pi$ and $\mathcal{T}$ for $T-1$ time-steps. We also define our value and Q-value functions as $V^{\pi}_t(s) = \mathop{{}\mathbb{E}}_{\tau \sim \pi | s_t=s}[\sum_{t'=t}^T r(s_{t'}, a_{t'})]$, $Q^{\pi}_t(s, a) = \mathop{{}\mathbb{E}}_{\tau \sim \pi | s_t=s, a_t=a}[\sum_{t'=t}^T r(s_{t'}, a_{t'})]$. We also define the advantage function as $A^{\pi}_t(s, a) = Q^{\pi}_t(s, a) - V^{\pi}_t(s)$. Lastly, let performance be $J(\pi) = \mathop{{}\mathbb{E}}_{\tau \sim \pi}[\sum_{t=1}^T r(s_t, a_t)]$.
Then,
\begin{definition}
We define the \textbf{imitation gap} as:
\vspace{-4pt}
\begin{equation} \label{eq:mm_obj}
J(\pi_E) - J(\pi)
\end{equation}
\end{definition}
\vspace{-4pt}
The goal of the learner is to minimize \eqref{eq:mm_obj} and \textit{close the imitation gap}. The unique challenge in imitation learning is that the reward function $r$ is unknown and the learner must rely on demonstrations from the expert $\pi_E$ to minimize the gap. A natural way to solve this problem is to empirically match \emph{all} moments of  $J(\pi_E)$. If all the moments are perfectly matched, regardless of the unknown reward function, the imitation gap must go to zero. We now delve into the various types of moments we can match.
\subsection{Moment Taxonomy}
Broadly speaking, a learner can focus on matching per-timestep \textit{reward} or over-the-horizon $Q$-\textit{value} moments of expert behavior. We use $\mathcal{F}_r: \mathcal{S} \times \mathcal{A} \rightarrow [-1, 1]$ to denote a class of reward functions, $\mathcal{F}_Q: \mathcal{S} \times \mathcal{A} \rightarrow [-T, T]$ to denote the set of $Q$ functions induced by sampling actions from some $\pi \in \Pi$, and $\mathcal{F}_{Q_E}: \mathcal{S} \times \mathcal{A} \rightarrow [-\overbar{Q}, \overbar{Q}]$ to denote the set of $Q$ functions induced by sampling actions from $\pi_E$. We assume all three function classes are closed under negation. Lastly, we refer to $H \in [0, 2\overbar{Q}]$ as the \textit{recoverability constant} of the problem and define it as follows:
 \begin{definition} A pair $(\pi_E, \mathcal{F}_{Q_E})$ of an expert policy and set of expert $Q$-functions is said to be $\mathbf{H}$-\textbf{recoverable} if $\forall s \in \mathcal{S}$, $\forall a \in \mathcal{A}$, $\forall f \in \mathcal{F}_{Q_E}$, $|f(s, a) - \mathop{{}\mathbb{E}}_{a' \sim \pi_E(s)}[f(s, a')]| < H$.
\end{definition}
$H$ is an upper bound on all possible advantages that could be obtained by the expert under a $Q$-function in $\mathcal{F}_{Q_E}$. Intuitively, $H$ tells us how many time-steps it takes the expert to recover from an arbitrary mistake. We defer a more in-depth discussion of the implications of this concept to Section \ref{sec:recover}. 

\noindent\textbf{Reward.} Matching reward moments entails minimizing the following expansion of the imitation gap:
\vspace{-4pt}
\begin{flalign*}
  \begin{aligned}
    \quad & J(\pi_E) - J(\pi) \\[-4pt]
    &= \mathop{{}\mathbb{E}}_{\tau \sim \pi_E}\sum\nolimits_{t=1}^T r(s_t, a_t) - \mathop{{}\mathbb{E}}_{\tau \sim \pi}\sum\nolimits_{t=1}^T r(s_t, a_t) \\[-4pt]
    &= \mathop{{}\mathbb{E}}_{\tau \sim \pi}\sum\nolimits_{t=1}^T -r(s_t, a_t) - \mathop{{}\mathbb{E}}_{\tau \sim \pi_E}\sum\nolimits_{t=1}^T -r(s_t, a_t) \\[-4pt]
    & \leq \sup_{f \in \mathcal{F}_r} \mathop{{}\mathbb{E}}_{\tau \sim \pi}\sum\nolimits_{t=1}^T f(s_t, a_t) - \mathop{{}\mathbb{E}}_{\tau \sim \pi_E}\sum\nolimits_{t=1}^T f(s_t, a_t)
  \end{aligned}&
  \vspace{-4pt}
\end{flalign*}
In the last step, we use the fact that $-r(s, a) \in \mathcal{F}_r$.  Crucially, reward moment-matching demands on-policy rollouts $\tau \sim \pi$ for the learner to calculate per-timestep divergences. 

Instead of matching moments of the reward function, we can consider  matching moments of the action-value function. We can apply the Performance Difference Lemma (PDL) to expand the imitation gap \eqref{eq:mm_obj} into either on-policy or off-policy expressions.

\noindent\textbf{Off-Policy $\mathbf{Q}$.} 
Starting from the PDL:
\vspace{-4pt}
\begin{flalign*}
  \begin{aligned}
&\quad  J(\pi_E) - J(\pi) \\[-4pt]
  &= \mathop{{}\mathbb{E}}_{\tau \sim \pi_E}[\sum\nolimits_{t=1}^T Q^{\pi}_t(s_t, a_t) - \mathop{{}\mathbb{E}}_{a \sim \pi(s_t)}[Q^{\pi}_t(s_t, a)]] \\[-4pt]
    & \leq \sup_{f \in \mathcal{F}_Q} \mathop{{}\mathbb{E}}_{\tau \sim \pi_E}[\sum\nolimits_{t=1}^T \mathop{{}\mathbb{E}}_{a \sim \pi(s_t)}[f(s_t, a)] - f(s_t, a_t)] 
  \end{aligned}&
  \vspace{-4pt}
\end{flalign*}
\begin{table}
\label{momentstable}
\vskip 0.15in
\begin{center}
\begin{small}
\begin{sc}
\begin{tabular}{lcccr}
\toprule
 Moment & Class & Env. & $\pi_E$ Queries\\
\midrule
 Reward & $\mathcal{F}_r: [-1, 1]$ & \cmark & \xmark \\ 
 Off-Policy $Q$ & $\mathcal{F}_Q: [-T, T]$ & \xmark & \xmark \\
 On-Policy $Q$ & $\mathcal{F}_{Q_E}: [-\overbar{Q}, \overbar{Q}]$ & \cmark  & \cmark \\
\bottomrule
\end{tabular}
\end{sc}
\end{small}
\end{center}
\vskip -0.1in
\caption{An overview of the requirements for the three classes of moment matching.}
\end{table}
In the last step, we use the fact that $Q^{\pi}_t(s, a) \in \mathcal{F}_Q$ for all $\pi \in \Pi$ and $r \in \mathcal{F}_r$. The above expression is off-policy -- it only requires a collected dataset of expert trajectories to be evaluated and minimized. In general though, $\mathcal{F}_Q$ can be a far more complex class than $\mathcal{F}_r$ because it has to capture both the dynamics of the MDP and the choices of \textit{any} policy.


\noindent\textbf{On-Policy $\mathbf{Q}$.} Expanding in the reverse direction:
\vspace{-4pt}
\begin{flalign*}
  \begin{aligned}
&\quad  J(\pi_E) - J(\pi) \\[-4pt]
  &= -\mathop{{}\mathbb{E}}_{\tau \sim \pi}[\sum\nolimits_{t=1}^T Q^{\pi_E}_t(s_t, a_t) - \mathop{{}\mathbb{E}}_{a \sim \pi_E(s_t)}[Q^{\pi_E}_t(s_t, a)]] \\[-4pt]
    & \leq \sup_{f \in \mathcal{F}_{Q_E}} \mathop{{}\mathbb{E}}_{\tau \sim \pi}[\sum\nolimits_{t=1}^T  f(s_t, a_t) - \mathop{{}\mathbb{E}}_{a \sim \pi_E(s_t)}[f(s_t, a)]] 
  \end{aligned}&
  \vspace{-4pt}
\end{flalign*}
In the last step, we use the fact that $Q^{\pi_E}_t(s, a) \in \mathcal{F}_{Q_E}$ for all $r \in \mathcal{F}_r$. In the realizable setting, $\pi_E \in \Pi$, $\mathcal{F}_{Q_E} \subseteq \mathcal{F}_{Q}$. While $\mathcal{F}_{Q_E}$ is a smaller class, to actually evaluate this expression, we require an \textit{interactive} expert that can tell us what action they would take in any state visited by the learner as well as on-policy samples from the learner's current policy $\tau \sim \pi$.

With this taxonomy in mind, we now turn our attention to deriving policy performance bounds. \footnote{See Sup. \ref{sec:other_moments} for mixed moments and an alternative $Q$-moment scheme that can be extended to the IL from observation alone setting.}

\subsection{Moment Matching Games}
A unifying perspective on the three moment matching variants can be achieved by viewing the learner as solving a game. More specifically, we consider variants of a two-player minimax game between a learner and a discriminator. The learner selects a policy $\pi \in \Pi$, where $\Pi \triangleq \{\pi: \mathcal{S} \rightarrow \Delta(\mathcal{A})\}$. We assume $\Pi$ is convex, compact and that $\pi_E \in \Pi$.\footnote{The full policy class satisfies all these assumptions.} The discriminator (adversarially) selects a function $f \in \mathcal{F}$, where $\mathcal{F} \triangleq  \{f: \mathcal{S} \times \mathcal{A} \rightarrow \mathbb{R}\}$. We assume that $\mathcal{F}$ is convex, compact, closed under negation, and finite dimensional.\footnote{Our results extend to infinite-dimensional Reproducing Kernel Hilbert Spaces.} Depending on the class of moments being matched, we assume that $\mathcal{F}$ is spanned by convex combinations of the elements of $\mathcal{F}_r / 2$, $\mathcal{F}_Q / 2T$, or $\mathcal{F}_{Q_E}/ 2T$.
 Lastly, we set the learner as the \emph{minimization player} and the discriminator as the \emph{maximization player}.

\begin{definition}
The \textbf{on-policy reward}, \textbf{off-policy $\mathbf{Q}$}, and \textbf{on-policy $\mathbf{Q}$} payoff functions are, respectively:
\begin{align}
U_1(\pi, f) =\frac{1}{T}(\mathop{{}\mathbb{E}}_{\tau \sim \pi}[\sum_{t=1}^T f(s_t, a_t)] - \mathop{{}\mathbb{E}}_{\tau \sim \pi_E}[\sum_{t=1}^T f(s_t, a_t)]) \nonumber \\
U_2(\pi, f) = \frac{1}{T}(\mathop{{}\mathbb{E}}_{\substack{\tau \sim \pi_E \\ a \sim \pi(s_t)}}[\sum_{t=1}^T f(s_t, a)] - \mathop{{}\mathbb{E}}_{\tau \sim \pi_E}[\sum_{t=1}^T f(s_t, a_t)])  \nonumber \\
U_3(\pi, f) = \frac{1}{T}(\mathop{{}\mathbb{E}}_{\tau \sim \pi}[\sum_{t=1}^T f(s_t, a_t)] - \mathop{{}\mathbb{E}}_{\substack{\tau \sim \pi \\ a \sim \pi_E(s_t)}}[\sum_{t=1}^T f(s_t, a)]) \nonumber
\end{align}
\end{definition}
When optimizing over the policy class $\Pi$ which contains $\pi_E$, we have a minimax value of 0: for $j \in \{1, 2, 3\}$,
\vspace{-4pt}
$$
\min_{\pi \in \Pi} \max_{f \in \mathcal{F}} U_j(\pi, f) = 0
$$
Furthermore, for certain representations of the policy,\footnote{One option is a mixture distribution over class $\Pi'$ where optimization is now performed over the mixture weights. This distribution can then be collapsed to a single policy \cite{syed2008apprenticeship}. A second option is to optimize over the causal polytope $P(\mathbf{A}^T||\mathbf{S}^T)$, as defined in \cite{ziebart2010modeling}.} strong duality holds: for $j\in \{1, 2, 3\}$,
$$
\min_{\pi \in \Pi} \max_{f \in \mathcal{F}} U_{j}(\pi, f) = \max_{f \in \mathcal{F}} \min_{\pi \in \Pi}  U_{j}(\pi, f)
$$
\vspace{-4pt}

We now study the properties that result from achieving an approximate equilibrium for each imitation game.

\section{From Approximate Equilibria to Bounded Regret}
\begin{table}[t]
\label{mmtable}
\vskip 0.15in
\begin{center}
\begin{small}
\begin{sc}
\begin{tabular}{lcccr}
\toprule
 Moment Matched & Upper Bound & Lower Bound \\
\midrule
 Reward & $O(\epsilon T)$ & $\Omega(\epsilon T)$ \\ 
 Off-Policy $Q$ & $O(\epsilon T^2)$ & $\Omega(\epsilon T^2)$ \\
 On-Policy $Q$ & $O(\epsilon HT)$ & $\Omega(\epsilon T)$ \\
\bottomrule
\end{tabular}
\end{sc}
\end{small}
\end{center}
\vskip -0.1in
\caption{An overview of the difference in bounds between the three types of moment matching. All bounds are on  imitation gap \eqref{eq:mm_obj}.}
\end{table}

A learner computing an equilibrium policy for any of the moment matching games will be imperfect due to many sources including restricted policy class, optimization error, or imperfect estimation of expert moments. More formally, in a game with payoff $U_j$, a pair $(\hat{\pi} \in \Pi, \hat{f} \in \mathcal{F})$ is a \textit{$\delta$-approximate equilibrium solution} if the following holds: \vspace{-4pt}
\begin{flalign*}
  \begin{aligned}
   \sup_{f \in \mathcal{F}} U_j(f, \hat{\pi}) - \frac{\delta}{2} \leq U_j(\hat{f}, \hat{\pi}) \leq \inf_{\pi \in \Pi} U_j(\hat{f}, \pi) + \frac{\delta}{2}
  \end{aligned}&
  \vspace{-4pt}
\end{flalign*}
We assume access to an \textit{algorithmic primitive} capable of finding such strategies:
\begin{definition} An \textbf{imitation game $\mathbf{\delta}$-oracle} $\mathbf{\Psi\{\delta\}(\cdot)}$ takes payoff function $U: \Pi \times \mathcal{F} \rightarrow [-k, k]$ and returns a $(k \delta)$-approximate equilibrium strategy for the policy player. \label{oracle:ilgame}
\vspace{-4pt}
\end{definition}
We now bound the imitation gap of solutions returned by such an oracle.


\subsection{Example MDPs}
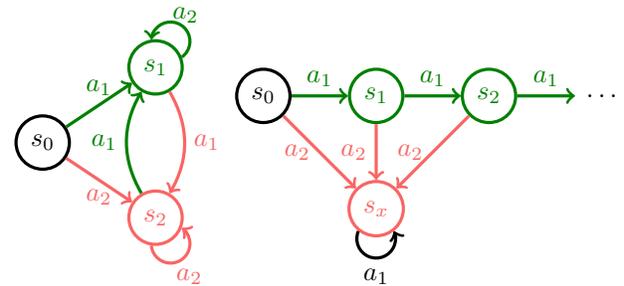
\begin{figure}[ht]
    \centering
    \begin{tikzpicture}[scale=1, transform shape]
    \node (a) [draw, very thick, circle] at (0.0, 0) {$s_0$};
    \node (b) [draw, very thick, circle, color=expert] at (1.5, 1) {$s_1$};
    \node (c) [draw, very thick, circle, color=error]  at (1.5, -1) {$s_2$};
    \path[->, color=error] (b) to[bend left] node[midway, right] {$a_1$} (c);
    \draw [->, very thick, color=error] (b) to[bend left] (c);
    \path[->, color=expert] (c) to[bend left] node[midway, left] {$a_1$} (b);
    \draw [->, very thick, color=expert] (c) to[bend left] (b);
    \path[->] (a) to node[midway, above, color=expert] {$a_1$} (b);
    \draw [->, very thick, color=expert] (a) to (b);
    \path[->] (a) to node[midway, below, color=error] {$a_2$} (c);
    \draw [->, very thick, color=error] (a) to (c);
    \node [circle, minimum size=0.5cm](d) at ([{shift=(60:0.4)}]b){};
    \coordinate (e) at (intersection 2 of b and d);
    \coordinate (f) at (intersection 1 of b and d);
    \tikzAngleOfLine(d)(f){\AngleStart}
    \tikzAngleOfLine(d)(e){\AngleEnd}
    \draw[very thick,->, color=expert]%
    let \p1 = ($ (d) - (f) $), \n2 = {veclen(\x1,\y1)}
    in
        (d) ++(60:0.4) node{$a_2$}
        (f) arc (\AngleStart-360:\AngleEnd:\n2);
        
    \node [circle, minimum size=0.5cm](g) at ([{shift=(300:0.4)}]c){};
    \coordinate (h) at (intersection 2 of c and g);
    \coordinate (i) at (intersection 1 of c and g);
    \tikzAngleOfLine(g)(i){\AngleStart}
    \tikzAngleOfLine(g)(h){\AngleEnd}
    \draw[very thick,->, color=error]%
    let \p1 = ($ (g) - (i) $), \n2 = {veclen(\x1,\y1)}
    in
        (g) ++(300:0.5) node{$a_2$}
        (i) arc (\AngleStart-360:\AngleEnd:\n2);
    \end{tikzpicture}
        \begin{tikzpicture}[scale=1, transform shape]
    \node (a) [draw, very thick, circle] at (0.0, 0) {$s_0$};
    \node (b) [draw, very thick, circle, color=expert] at (1.5, 0) {$s_1$};
    \node (c) [draw, very thick, circle, color=expert]  at (3, 0) {$s_2$};
    \node (d) [circle]  at (4.5, 0) {$\ldots$};
    \node (e) [draw, very thick, circle, color=error]  at (1.5, -1.5) {$s_x$};
    \path[->,] (a) to node[midway, above, color=expert] {$a_1$} (b);
    \draw [->, very thick, color=expert] (a) to (b);
    \path[->] (b) to node[midway, above, color=expert] {$a_1$} (c);
    \draw [->, very thick, color=expert] (b) to (c);
    \path[->] (c) to node[midway, above, color=expert] {$a_1$} (d);
    \draw [->, very thick, color=expert] (c) to (d);
    \path[->] (a) to node[midway, left, color=error] {$a_2$} (e);
    \draw [->, very thick, color=error] (a) to (e);
    \path[->] (b) to node[midway, left, color=error] {$a_2$} (e);
    \draw [->, very thick, color=error] (b) to (e);
    \path[->] (c) to node[midway, left, color=error] {$a_2$} (e);
    \draw [->, very thick, color=error] (c) to (e);

    \node [circle, minimum size=0.5cm](g) at ([{shift=(270:0.4)}]e){};
    \coordinate (h) at (intersection 2 of e and g);
    \coordinate (i) at (intersection 1 of e and g);
    \tikzAngleOfLine(g)(i){\AngleStart}
    \tikzAngleOfLine(g)(h){\AngleEnd}
    \draw[very thick,->]%
    let \p1 = ($ (g) - (i) $), \n2 = {veclen(\x1,\y1)}
    in
        (g) ++(270:0.5) node{$a_1$}
        (i) arc (\AngleStart-360:\AngleEnd:\n2);
    \end{tikzpicture}
    \caption{Left: Borrowed from \cite{ross2011reduction}, the goal of \textsc{Loop} is to spend time in $s_1$. Right: a folklore MDP \textsc{Cliff}, where the goal is to not ``fall off the cliff" and end up in $s_x$ evermore.}
    \label{fig:mdps}\vspace{-1em}
\end{figure}

For use in our analysis, we first introduce two MDPs, \textsc{Loop} and \textsc{Cliff}. As seen in \figref{fig:mdps}, \textsc{Loop} is an MDP where a learner can enter a state where it has seen no expert demonstrations ($s_2$) and make errors for the rest of the horizon. \textsc{Cliff} is an MDP where a single mistake can result in the learner being stuck in an absorbing state.

\subsection{Reward Moment Performance Bounds}
Let us first consider the reward moment-matching game. 

\begin{lemma}{\textbf{Reward Upper Bound:}} If $\mathcal{F}_r$ spans $\mathcal{F}$, then for all MDPs, $\pi_E$, and $\pi \leftarrow \Psi\{\epsilon\}(U_1)$, $J(\pi_E) - J(\pi) \leq O(\epsilon T)$.
\end{lemma}
\begin{pf} We start by expanding the imitation gap:
\vspace{-4pt}
\begin{flalign*}
  \begin{aligned}
    \quad & J(\pi_E) - J(\pi) \\[-4pt]
    & \leq \sup_{f \in \mathcal{F}_r} \mathop{{}\mathbb{E}}_{\tau \sim \pi}\sum\nolimits_{t=1}^T f(s_t, a_t) - \mathop{{}\mathbb{E}}_{\tau \sim \pi_E}\sum\nolimits_{t=1}^T f(s_t, a_t) \\[-4pt]
    & \leq \sup_{f \in \mathcal{F}} \mathop{{}\mathbb{E}}_{\tau \sim \pi}\sum\nolimits_{t=1}^T 2f(s_t, a_t) - \mathop{{}\mathbb{E}}_{\tau \sim \pi_E}\sum\nolimits_{t=1}^T 2f(s_t, a_t) \\[-4pt]
    & = 2T \sup_{f \in \mathcal{F}} U_1(\pi, f) \leq 2T \epsilon
  \end{aligned}&
  \vspace{-4pt}
\end{flalign*}
The first line follows from the closure of $\mathcal{F}_r$ under negation. The last line follows from the definition of an $\epsilon$-approximate equilibrium.
\end{pf}
In words, this bound means that in the worst case, we have an imitation gap that is $O(\epsilon T)$ rather than an imitation gap that compounds quadratically in time.\textbf{}
\begin{lemma}{\textbf{Reward Lower Bound:}}  There exists an MDP, $\pi_E$, and $\pi \leftarrow \Psi\{\epsilon\}(U_1)$ such that $J(\pi_E) - J(\pi) \geq \Omega(\epsilon T)$.
\end{lemma}


\begin{pf}
Consider \textsc{Cliff} with a reward function composed of two indicators: $r(s, a) = -\mathds{1}_{s_x} -\mathds{1}_{a_2}$ and a perfect expert that never takes $a_2$. If with probability $\epsilon$ the learner's policy takes action $a_2$ only in $s_0$, the optimal discriminator would not only be able to penalize the learner for taking $a_2$ but also for the next $T-1$ timesteps for being in $s_x$. Together, this would lead to an average cost of $\epsilon$ per timestep. Under $r$, this would make the learner $\epsilon T$ worse than the expert, giving us $J(\pi_E) - J(\pi) = \epsilon T \geq \Omega(\epsilon T)$.
\end{pf}

Notably, both of these bounds are purely a function of the \textit{game}, not the policy search algorithm and therefore apply for \textit{all} algorithms that can be written in the form of a reward moment-matching imitation game. Our bounds do not depend on the size of the state space and therefore apply to continuous spaces, unlike those presented in \cite{rajaraman2020fundamental}. Several recently proposed algorithms \cite{ho2016generative, brantley2020disagreement-regularized, ALICE, Yang_Vereshchaka_Zhou_Chen_Dong_2020} including GAIL and SQIL can be understood as also solving this or a related game.

\subsection{Off-$\mathbf{Q}$ Moment Performance Bounds}
We contrast the preceding guarantees with those based on matching off-$Q$ moments.
\begin{lemma}{\textbf{Off-$\mathbf{Q}$ Upper Bound:}} If $\mathcal{F}_Q$ spans $\mathcal{F}$, then for all MDPs, $\pi_E$, and $\pi \leftarrow \Psi\{\epsilon\}(U_2)$, $J(\pi_E) - J(\pi) \leq O(\epsilon T^2)$.
\end{lemma}
\begin{pf}
Starting from the PDL:
\vspace{-4pt}
\begin{flalign*}
  \begin{aligned}
&\quad  J(\pi_E) - J(\pi) \\[-4pt]
    & \leq \sup_{f \in \mathcal{F_Q}} \mathop{{}\mathbb{E}}_{\tau \sim \pi_E}[\sum\nolimits_{t=1}^T \mathop{{}\mathbb{E}}_{a \sim \pi(s_t)}[f(s_t, a)] - f(s_t, a_t)] \\[-4pt]
    & \leq \sup_{f \in \mathcal{F}} \mathop{{}\mathbb{E}}_{\tau \sim \pi_E}[\sum\nolimits_{t=1}^T \mathop{{}\mathbb{E}}_{a \sim \pi(s_t)}[2Tf(s_t, a)] - 2Tf(s_t, a_t)] \\ 
    & = 2T^2 \sup_{f \in \mathcal{F}} U_2 (\pi, f) \leq 2\epsilon T^2
  \end{aligned}&
  \vspace{-4pt}
\end{flalign*}
The $T$ in the second to last line comes from the fact that $\mathcal{F}_Q / 2T \subseteq \mathcal{F}$. Thus, a policy $\pi$ returned by $\Psi\{\epsilon\}(U_2)$ must satisfy $J(\pi_E) - J(\pi) \leq O(\epsilon T^2)$ -- that is, it can do up to $O(\epsilon T^2)$ worse than the expert.
\end{pf}
\begin{lemma}{\textbf{Off-$\mathbf{Q}$ Lower Bound:}} There exists an MDP, $\pi_E$, and $\pi \leftarrow \Psi\{\epsilon\}(U_2)$ such that $J(\pi_E) - J(\pi) \geq \Omega(\epsilon T^2)$.
\end{lemma}
\begin{pf}
Once again, consider \textsc{Cliff}. If the learner policy instead takes $a_2$ with probability $\epsilon T$ in $s_0$, the optimal discriminator would be able to penalize the learner up to $\epsilon T$ for that timestep and $\epsilon$ on average. However, on rollouts, the learner would have an $\epsilon T$ chance of paying a cost of 1 for the rest of the horizon, leading to a lower bound of $J(\pi_E) - J(\pi) = \epsilon T^2 \geq \Omega(\epsilon T^2)$.
\end{pf}

These bounds apply for \textit{all} algorithms that can be written in the form of an off-$Q$ imitation game, including behavioral cloning \cite{pomerleau1989alvinn} and ValueDICE \cite{kostrikov2019imitation}.
\subsection{On-$\mathbf{Q}$ Moment Performance Bounds}
We now derive performance bounds for on-$Q$ algorithms with interactive experts.
\begin{lemma}{\textbf{On-$\mathbf{Q}$ Upper Bound:}} If $\mathcal{F}_{Q_E}$ spans $\mathcal{F}$, then for all MDPs with $H$-recoverable $(\mathcal{F}_{Q_E}, \pi_E)$, and $\pi \leftarrow \Psi\{\epsilon\}(U_3)$, $J(\pi_E) - J(\pi) \leq O(\epsilon HT)$.
\end{lemma}
\begin{pf}
Starting from the PDL:
  \begin{align}
&\quad  J(\pi_E) - J(\pi) \nonumber\\[-4pt]
    & \leq \sup_{f \in \mathcal{F}_{Q_E}} \mathop{{}\mathbb{E}}_{\tau \sim \pi}[\sum\nolimits_{t=1}^T f(s_t, a_t) - \mathop{{}\mathbb{E}}_{a \sim \pi_E(s_t)}[f(s_t, a_t)] ]  \nonumber \\[-4pt]
    & \leq \sup_{f \in \mathcal{F}} \mathop{{}\mathbb{E}}_{\tau \sim \pi_E}[\sum\nolimits_{t=1}^T 2T f(s_t, a_t) - \mathop{{}\mathbb{E}}_{a \sim \pi_E(s_t)}[2Tf(s_t, a)]] \nonumber \\ 
    & = 2T^2 \sup_{f \in \mathcal{F}} U_3 (\pi, f) \leq (\epsilon\frac{H}{2T}) 2T^2 = \epsilon H T\nonumber
  \end{align}
As before, the $T$ in the second to last line comes from the fact that $\mathcal{F}_{Q_E} / 2T \subseteq \mathcal{F}$. The $H / 2T$ factor comes from the scale of the payoff. Thus, a policy $\pi$ returned by $\Psi\{\epsilon\}(U_3)$ must satisfy $J(\pi_E) - J(\pi) \leq O(\epsilon H T)$ -- that is, it can do up to $O(\epsilon H T)$ worse than the expert.
\end{pf}
\begin{lemma}{\textbf{On-$\mathbf{Q}$ Lower Bound:}} There exists an MDP, $\pi_E$, and $\pi \leftarrow \Psi\{\epsilon\}(U_3)$ such that $J(\pi_E) - J(\pi) \geq \Omega(\epsilon T)$.
\label{bound:onq_low}
\end{lemma}
\begin{pf}
The proof of the \textit{reward lower bound} holds verbatim because every policy, including the previously considered $\pi_E$ will be stuck in $s_x$ after it falls in.
\end{pf}

As before, these bounds apply for all algorithms that can be written in the form of an on-$Q$ imitation game, including DAgger \cite{ross2011reduction} and AggreVaTe \cite{ross2014reinforcement}. For example, in the bounds for AggreVaTe, $Q_{max}$ is equivalent to the recoverability constant $H$.

\subsection{Recoverability in Imitation Learning}
\label{sec:recover}

The bounds we presented above beg the question of when on-$Q$ moment matching has error properties similar to those of reward moment matching versus those of off-$Q$ moment matching. Recoverability allows us to cleanly answer this question and others. We begin by providing more intuition for said concept.

Concretely, in \figref{fig:mdps}, \textsc{Loop} is $1$-recoverable for the expert policy that always moves towards $s_1$. \textsc{Cliff} is not $H$-recoverable for any $H<T$ if the expert never ends up in $s_x$. A sufficient condition for $H$-recoverability is that the state occupancy distribution that results from taking an arbitrary action and then $H-1$ actions according to $\pi_E$ is the same as that of taking $H$ actions according to $\pi_E$. We emphasize that recoverability is a property of the \textit{set of moments} matched and the expert, not just of the expert, as has been previously considered \cite{Pan-IJRR-19}.

\noindent \textbf{Bound Clarification.} Our previously derived upper bound for on-$Q$ moment matching ($J(\pi_E) - J(\pi) \leq O(\epsilon HT)$) tells us that for $O(1)$-recoverable MDPs, on-$Q$ moment matching behaves like reward moment matching while for $O(T)$-recoverable MDPs, it instead behaves like off-$Q$ moment matching and has an $O(\epsilon T^2)$ upper bound. Thus, $O(1)$-recoverability is in a certain sense necessary for achieving $O(\epsilon T)$ error with on-$Q$ moment matching.

Another perspective on recoverability is that it helps us delineate problems where compounding errors are hard to avoid for both on-$Q$ and reward moment matching. Let $l(s)=\sum_{a' \in \mathcal{A}}|\mathop{{}\mathbb{E}}_{a \sim \pi_E(s)}[\mathds{1}_{a'}(a)] - \mathop{{}\mathbb{E}}_{a \sim \pi(s)}[\mathds{1}_{a'}(a)]|$ be the classification error of a state $s$. We prove the following lemma in Supplementary Material \ref{proof:nonreclb}:
\begin{lemma}
\label{thm:nonreclb} Let $\kappa > 0$. There exists a ($\pi_E$, $\{r, -r\}$) pair that for any $H < T$ is not $H$-\textit{recoverable} such that $l(s) = \kappa$ on any state leads to $J(\pi_E) - J(\pi) \geq \Omega(\kappa T^2)$.
\end{lemma}
Because some states might not appear on expert rollouts, evaluating $l(s)$ can require an interactive expert. However, even with this strong form of feedback, a classification error of $\kappa$ on a single state can lead to $\Omega(\kappa T^2)$ imitation gap for $O(T)$-recoverable MDPs. This lemma also implies that in such an MDP, achieving $O(\kappa T)$ imitation gap via on-policy moment matching would require the learner to have a classification error $\propto \kappa / T$, or to make vanishingly rare errors as we increase the horizon. We note that this does not conflict with our previously stated bounds but reveals that achieving a moment-matching error of (time-independent) $\epsilon$ might require achieving a classification error that scales inversely with time for $O(T)$-recoverable MDPs. Practically, this can be rather challenging. Thus, neither on-$Q$ nor reward moment matching is a silver bullet for getting $O(\epsilon T)$ error for $O(T)$-recoverable problems.

\section{Finding Equilibria: Idealized Algorithms}
\label{sec:equib}
We now provide reduction-based methods for computing (approximate) equilibria for these moment matching games which can be seen as blueprints for constructing our previously described oracle (Def. \ref{oracle:ilgame}). We study, in particular, finite state problems and a complete policy class. We analyze an approach to equilibria finding where an outer player follows a \textit{no-regret} strategy and the inner player follow a (modified) \textit{best response} strategy, by which we can efficiently find policies with strong performance guarantees. 

\subsection{Preliminaries}
An \textit{efficient no-regret algorithm} over a class $\mathcal{X}$ produces iterates  $x^1, \dots, x^{H} \in \mathcal{X}$ that satisfy the following property for any sequence of loss functions $l^1, \dots, l^{H}$:
\vspace{-4pt}
\begin{flalign*}
  \begin{aligned}
 \text{Regret}(H) = \sum\nolimits_t^{H} l^t(x^t) - \min_{x \in \mathcal{X}} \sum\nolimits_t^{H} l^t(x) \leq \beta_{\mathcal{X}}(H) 
  \end{aligned}&
  \vspace{-4pt}
\end{flalign*}
where $\beta_{\mathcal{X}}(H)  / {H} \leq \epsilon$ holds for $H$ that are  $O(\textrm{poly}(\frac{1}{\epsilon}))$.
\subsection{Theoretical Guarantees}

We are interested in obtaining a policy efficiently that is a near-equilibrium solution to the game. We consider two general strategies to do so:

\noindent\textbf{Primal.} We execute a no-regret algorithm on the policy representation, while a maximization oracle over the space $\mathcal{F}$ computes the best response to those policies.

\noindent\textbf{Dual.} We execute a no-regret algorithm on the space $\mathcal{F}$, while a minimization oracle over policies computes \textit{entropy regularized} best response policies.

The asymmetry in the above is driven by the need to recover the equilibrium strategy for the policy player and the fact that a dual approach on the original, unregularized objective $U_j(\cdot,f)$ will typical not converge to a single policy but rather shift rapidly as $f$ changes.\footnote{The average over iterations of the policies generated in an unregularized dual will also be near-equilibrium but can be inconvenient. Entropy regularization provides a convenient way to extract a single policy and meshes well with empirical practice.}

By choosing the the policy representation to be a causally conditioned probability distribution over actions,
$P(\mathbf{A}^T||\mathbf{S}^T) = \prod_{t=1}^T P(A_t|\mathbf{S}_{1:t}, \mathbf{A}_{1:t-1})$, we find each of the imitation games is \textit{bilinear} in both policy and discriminator $f$ and strongly dual .\footnote{Following \cite{ziebart2010modeling} we can represent the policy as an element of the causal conditioned polytope and regularize with the causal entropy $H(P(\mathbf{A^T}||\mathbf{S^T}))$ to denote the causal Shannon entropy of a policy. An equivalent result can be proved for optimizing over \textit{occupancy measures} \cite{ho2016generative}.} Thus, we can efficiently compute a near-equilibrium assuming access to the optimization oracles in either \textit{primal} or \textit{dual} above:
\clearpage
\begin{theorem}
\label{thm:master}
Given access to the no-regret and maximization oracles in either \textbf{primal} or \textbf{dual} above, for all three imitation games 
we are able to compute a $\delta$-approximate equilibrium strategy for the policy player in ${\normalfont poly}(\frac{1}{\delta}, T, \ln{|\mathcal{S}|}, \ln{|\mathcal{A}|})$ iterations of the outer player optimization.
\end{theorem}
This result relies on a general lemma that establishes we can recover the equilibrium inner player by appropriately entropy regularizing that inner player's decisions. We prove both in Sup. \ref{sec:proofs}.

By substituting $\delta = \epsilon$ in Theorem \ref{thm:master} and combining with our previously derived bounds, we can cleanly show that these theorems can also be viewed as certificates of policy performance. Thus, our framework enables us to efficiently \textit{close the imitation gap}.

\section{Practical Moment Matching Algorithms}

We now present three practical algorithms that match reward moments (\texttt{AdRIL}), off-$Q$ moments (\texttt{AdVIL}), or on-$Q$ moments (\texttt{DAeQuIL}). They are specific, implementable versions of our preceding abstract procedures. At their core, all three algorithms optimize an IPM. IPMs are a distance between two probability distributions \cite{muller1997predicting}:
\vspace{-4pt}
$$ \sup_{f \in \mathcal{F}} \mathop{{}\mathbb{E}}_{x \sim P_1}[f(x)] - \mathop{{}\mathbb{E}}_{x \sim P_2}[f(x)]$$
Plugging in the learner and expert trajectory distributions, we end up with our IPM-based objective:
\vspace{-4pt}
\begin{equation} \label{eq:ipm}
\sup_{f \in \mathcal{F}} \sum_{t = 1}^T \mathop{{}\mathbb{E}}_{\tau \sim \pi}[f(s_t, a_t)] - \mathop{{}\mathbb{E}}_{\tau \sim \pi_E}[f(s_t, a_t)]
\vspace{-4pt}
\end{equation}
As noted previously, this objective is equivalent to and inherits the strong guarantees of moment matching and allows our analysis to easily transfer.



\subsection{\texttt{AdVIL}: Adversarial Value-moment Imitation Learning}

\begin{algorithm}[t]
\begin{algorithmic}
\STATE {\bfseries Input:} Expert demonstrations $\mathcal{D}_E$, Policy class $\Pi$, Discriminator class $\mathcal{F}$, Performance threshold $\delta$, Learning rates $\eta_f > \eta_{\pi}$
\STATE {\bfseries Output:} Trained policy $\pi$
 \STATE Set $\pi \in \Pi$, $f \in \mathcal{F}$, $L(\pi, f) = 2 \delta$
 \WHILE{$L(\pi, f) > \delta$}
 \STATE $L(\pi, f) = \mathop{{}\mathbb{E}}_{(s, a) \sim \mathcal{D}_E}[\mathop{{}\mathbb{E}}_{x \sim \pi(s)}[ f(s, x)]-  f(s, a)]$
 \STATE $f \leftarrow f + \eta_f \nabla_f L(\pi, f) $
 \STATE $\pi \leftarrow \pi -  \eta_{\pi} \nabla_{\pi} L(\pi, f) $
 \ENDWHILE
\end{algorithmic}
\caption{\texttt{AdVIL}}
\label{alg:adril}
\end{algorithm}
We can transform our IPM-based objective \eqref{eq:ipm} into an expression only over $(s, a)$ pairs from expert data to fit into the off-$Q$ moment matching framework by performing a series of substitutions. We refer interested readers to Supplementary Material \ref{advil_derivation}. We arrive at the following expression:
\vspace{-4pt}
\begin{equation} \label{eq:advil_obj}
\sup_{v \in \mathcal{F}} 
\mathop{{}\mathbb{E}}_{\tau \sim \pi_E}[   
\sum_{t = 1}^T  \mathop{{}\mathbb{E}}_{a \sim \pi(s_{t})}[v(s_{t}, a)]
- v(s_t, a_t)]
\vspace{-4pt}
\end{equation}
Intuitively, by minimizing \eqref{eq:advil_obj} over policies $\pi \in \Pi$, one is driving down the extra cost the learner has over the expert. We term this approach \textit{Adversarial Value-moment Imitation Learning} (\texttt{AdVIL}) because if we view $\pi$ as optimizing cumulative reward, $-v$ could be viewed as a value function. We set the learning rate for $f$ to be greater than that for $\pi$, making \texttt{AdVIL} a primal algorithm.

Practically, \texttt{AdVIL} bears similarity to the Wasserstein GAN (WGAN) framework \cite{arjovsky2017wasserstein}, though we consider an IPM rather than the more restricted Wasserstein distances. However, the overlap is enough that techniques from the WGAN literature including gradient penalties on the discriminator \cite{gulrajani2017improved} and orthogonal regularization of the policy player \cite{brock2018large} help.

\subsection{\texttt{AdRIL}: Adversarial Reward-moment Imitation Learning}
\begin{algorithm}[t]
\begin{algorithmic}
\STATE {\bfseries Input:} Expert demonstrations $\mathcal{D}_E$, Policy class $\Pi$, Dynamics $\mathcal{T}$, Kernel $K$, Performance threshold $\delta$ 
\STATE {\bfseries Output:} Trained policy $\pi$
 \STATE Set $\pi \in \Pi$, $f = 0 $ , $\mathcal{D}_{\pi} = \{\}$, $\mathcal{D}' = \{\}$, $L(\pi, f) = 2 \delta$
 \WHILE{$L(\pi, f) > \delta$}
 \STATE $f \leftarrow \mathop{{}\mathbb{E}}_{\tau \sim D_{\pi}}[\sum_t K(sa, \cdot)] - \mathop{{}\mathbb{E}}_{\tau \sim D_E}[\sum_t K(sa, \cdot)]$
 \STATE $\pi, \mathcal{D}' \leftarrow \texttt{MaxEntRL}(\texttt{T}=\mathcal{T}, \texttt{r}=-f)$
 \STATE $\mathcal{D}_{\pi} \leftarrow \mathcal{D}_{\pi} \cup \mathcal{D}'$
 \STATE $L(\pi, f) = \mathop{{}\mathbb{E}}_{\tau \sim D'}[\sum_t f(s, a)] - \mathop{{}\mathbb{E}}_{\tau \sim D_E}[\sum_t f(s, a)]$ 
 \ENDWHILE
\end{algorithmic}
\caption{\texttt{AdRIL}}
\label{alg:adril}
\end{algorithm}
\begin{figure*}[ht]
  \includegraphics[width=\textwidth]{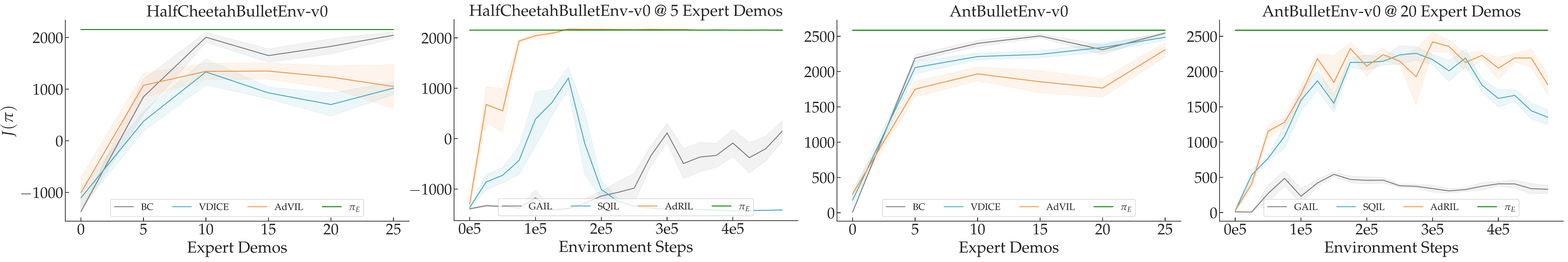}
  \caption{Our proposed methods (in orange) are able to match or out-perform the baselines across a variety of continuous control tasks. $J(\pi)$s are averaged across 10 evaluation episodes. Standard errors are across 5 runs of each algorithm. \label{fig:gokuls_tears}}
\end{figure*}
We now present our dual reward moment matching algorithm and refer interested readers to Supplementary Material \ref{adril_derivation} for the full derivation. In brief, we solve for the discriminator in closed form via functional gradient descent in Reproducing Kernel Hilbert Space (RKHS) and have the policy player compute a best response via maximum entropy reinforcement learning. Let $\overbar{\mathcal{D}_k}$ denote the aggregated dataset of policy rollouts. Assuming a constant number of training steps at each iteration and averaging functional gradients over $k$ iterations of the algorithm, we get the cost function for the policy and round $k$:
\vspace{-4pt}
\begin{flalign*}
  \begin{aligned}
 \sum_{t = 1}^T  \frac{1}{|\overbar{\mathcal{D}_k}|}\sum_{\tau}^{\overbar{\mathcal{D}_k}}K([s_t, a_t], \cdot) - \frac{1}{|\mathcal{D}_E|} \sum_{\tau}^{\mathcal{D}_E}K([s_t, a_t], \cdot) \label{eq:adril}
  \end{aligned}
  \vspace{-4pt}
\end{flalign*}
For an indicator kernel and under the assumption we never see the same state twice, this is equivalent to maximizing a reward function that is $1$ at each expert datapoint, $\propto \frac{-1}{k}$ along previous rollouts that do not perfectly match the expert, and $0$ everywhere else. We term this approach \textit{Adversarial Reward-moment Imitation Learning} (\texttt{AdRIL}). 

We note that under these assumption, our objective resembles that of SQIL \cite{reddy2019sqil}. SQIL can be seen as a degenerate case of \texttt{AdRIL} that never updates the discriminator function. This oddity removes solution quality guarantees while introducing the need for early stopping \cite{arenz2020nonadversarial}.

\subsection{\texttt{DAeQuIL}: DAgger-esque Qu-moment Imitation Learning}
We present the natural extension of DAgger \cite{ross2011reduction} to the space of moments: \texttt{DAeQuIL} (DAgger-esque Qu-moment Imitation Learning) in Algorithm \ref{alg:daequil}. Like DAgger, one can view \texttt{DAeQuIL} as a primal algorithm that uses Follow the Regularized Leader as the no-regret algorithm for the policy player. Per-round losses are adversarially chosen though. It is a subtle point but as written, \texttt{DAeQuIL} is technically not solving the on-$Q$ game directly because it optimizes over a history of learner state distributions instead of the current learner's state distribution. However, it retains strong performance guarantees -- see Sup. \ref{daequil_derivation} for more.
\begin{algorithm}[h]
\begin{algorithmic}
\STATE {\bfseries Input:} Queryable expert $\pi_E$, Policy class $\Pi$, Discriminator class $\mathcal{F}$, Performance threshold $\delta$, Behavioral cloning loss $\ell_{BC}: \Pi \rightarrow \mathbb{R}$, Strongly convex fn $R: \Pi \rightarrow \mathbb{R}$
\STATE {\bfseries Output:} Trained policy $\pi$
\STATE Optimize: $\pi \leftarrow \arg\min_{\pi' \in \Pi} \ell_{BC}(\pi')$.
\STATE Set $L(\pi) = 2\delta$, $\mathcal{D} = []$, $F = []$, $t=1$
 \WHILE{$L(\pi) > \delta$}
 \STATE Rollout $\pi$ to generate $\mathcal{D}_{\pi} \leftarrow [(s, a), \dots]$.
 \STATE Relabel $\mathcal{D}_{\pi}$ to $\mathcal{D}_E \leftarrow [(s, a') | a' \sim \pi_E(s) ,\, \forall s \in \mathcal{D}_{\pi} ]$
 \STATE $L(f) = \mathop{{}\mathbb{E}}_{(s, a) \sim \mathcal{D}_{\pi}}[f(s, a)] - \mathop{{}\mathbb{E}}_{(s, a) \sim \mathcal{D}_E}[f(s, a)]$
 \STATE Append: $F \leftarrow F \cup [\arg\max_{f' \in \mathcal{F}} L(f')]$.
 \STATE Append: $\mathcal{D} \leftarrow \mathcal{D} \cup [(s, t) \, | \, \forall s \in \mathcal{D}_{\pi}]$.
 \STATE $L(\pi) = \mathbb{E}_{(s, t) \in \mathcal{D}}[F[t](s, \pi(s))] + \ell_{BC}(\pi) + R(\pi)$
 \STATE Optimize: $\pi \leftarrow \arg\min_{\pi' \in \Pi} L(\pi')$.
 \STATE $t \leftarrow t + 1$
 \ENDWHILE
\end{algorithmic}
\caption{\texttt{DAeQuIL}}
\label{alg:daequil}
\end{algorithm}

\section{Experiments}
\begin{figure}[ht]
    \centering
    \includegraphics[width=3\columnwidth/10]{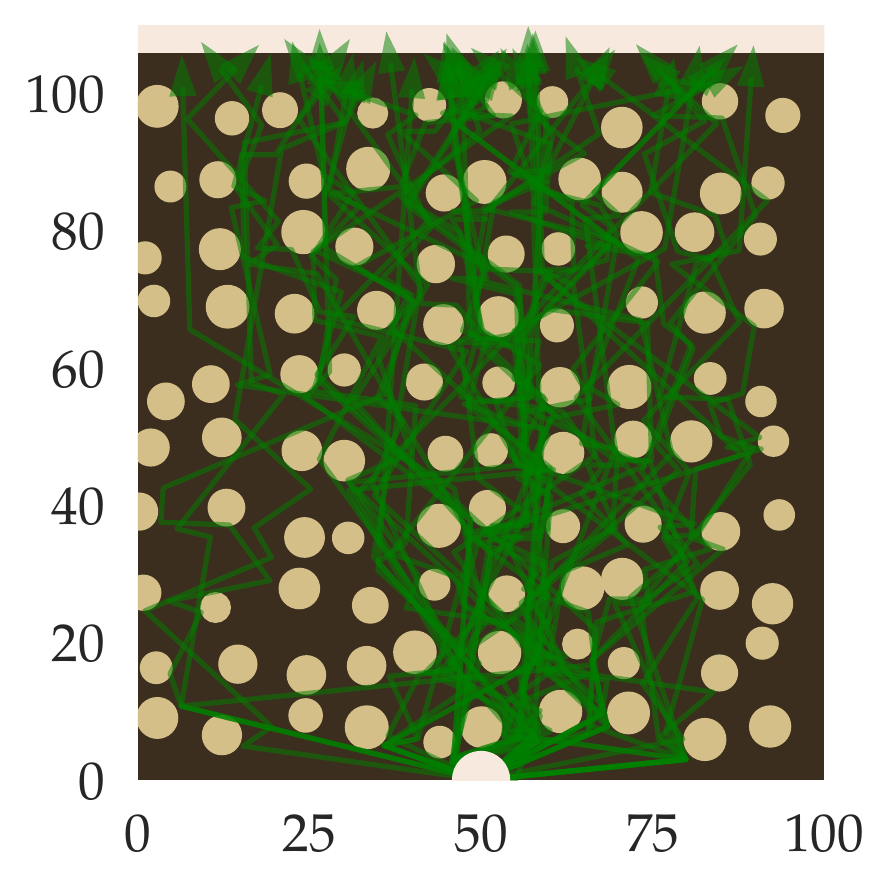}
    \includegraphics[width=3\columnwidth/10]{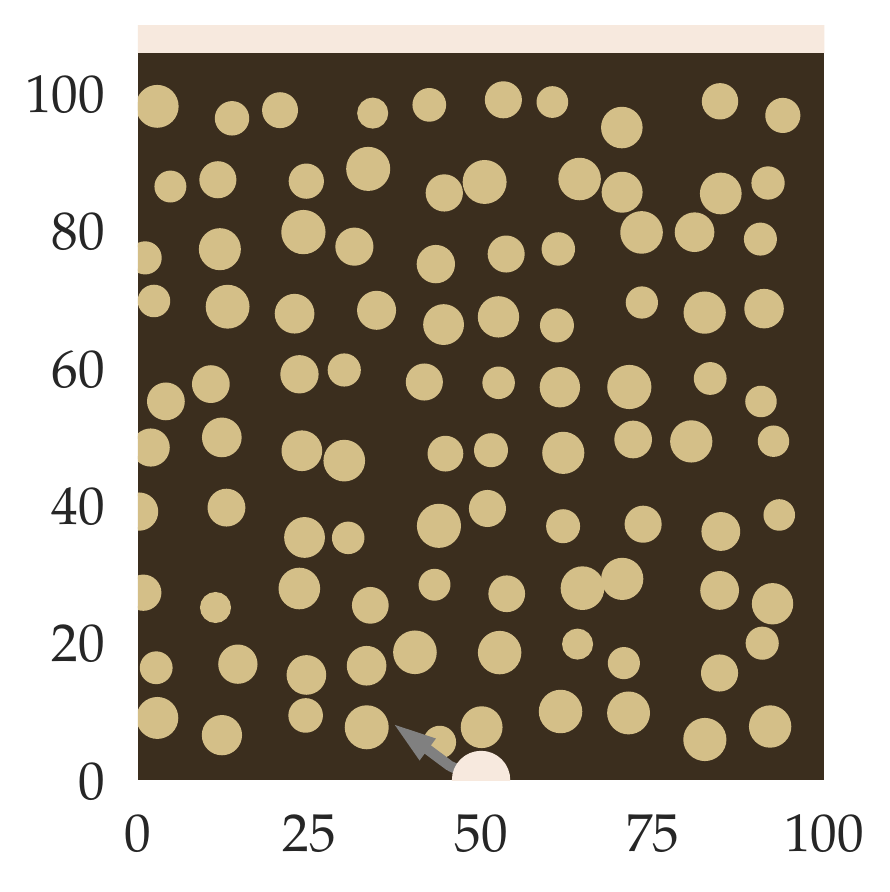}
    \includegraphics[width=3\columnwidth/10]{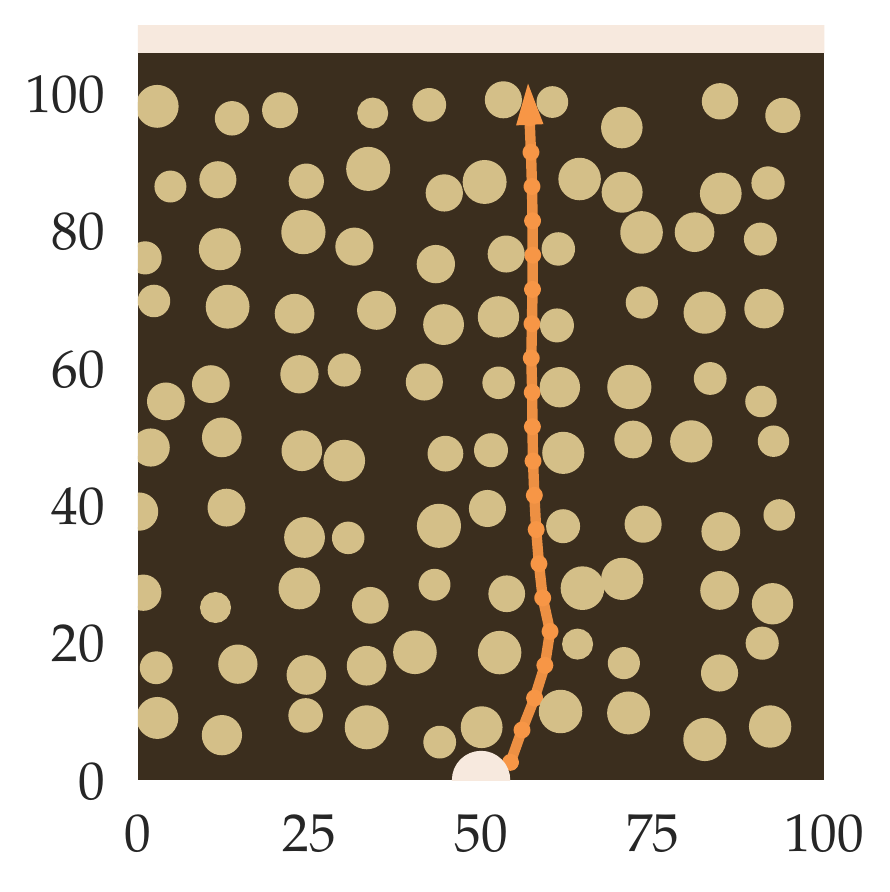}
    \caption{\textbf{Left:} The expert demonstrates many feasible trajectories, causing a learner that attempts to just match the mean action to crash directly into the tree the expert was avoiding. \textbf{Center:} On-policy corrections do not help DAgger as it still tries to match the mean action and crashes into the first tree it encounters. \textbf{Right:} \texttt{DAeQuIL}, when run with moments that allow the learner to focus on swerving out of the way of trees, regardless of direction, is able to produce policies that successfully navigate through the forest.}
    \label{fig:forest}
\end{figure}

We test our algorithms against several baselines on several higher-dimensional continuous control tasks from the PyBullet suite \cite{coumans2019}. We measure the performance of off-$Q$ algorithms as a function of the amount of data provided with a fixed maximum computational budget and of reward moment-matching algorithms as a function of the amount of environment interactions. We see from \figref{fig:gokuls_tears} that \texttt{AdVIL} can match the performance of ValueDICE and Behavioral Cloning across most tasks. \texttt{AdRIL} performs better than GAIL across all environments and does not exhibit the catastrophic collapse in performance SQIL does on the tested environments. On both environments, behavioral cloning is able to recover the optimal policy with enough data, indicating there is little covariate shift \cite{ALICE}. However, on HalfCheetah, we see \texttt{AdRIL} recover a strong policy with far less data than it takes Behavioral Cloning to, showcasing the potential benefit of the learner observing the consequences of their own actions. We refer readers to Sup. \ref{params} for a description of our hyparameters and setup. Notably, \texttt{AdVIL} is able to converge reliably to a policy of the same quality as that found by ValueDICE with an order of magnitude less compute. As seen in \figref{fig:forest}, \texttt{DAeQuIL} is able to significantly out-perform DAgger on a toy UAV navigation task -- see Sup. \ref{sec:onq_exp} for full information. We release our code at \textbf{\texttt{\url{https://github.com/gkswamy98/pillbox}}}.

\section{Discussion} 
\label{sec:discussion}
\subsection{A Unifying View of Moment Matching IL}
We present a cohesive perspective of moment matching in imitation learning in Table \ref{onetabletorulethemall}. We note that reward moment-matching dual algorithms have been a repeated success story in imitation learning but that there has been comparatively less work done in off-$Q$ and on-$Q$ dual algorithms.

\begin{table}[h]
\begin{center}
\begin{small}
\begin{sc}
\setlength{\tabcolsep}{2pt}
\begin{tabular}{lccccr}
\toprule
 Moment & Primal & Dual\\
\midrule
Off-$Q$ & VDICE, \textbf{AdVIL} &  \xmark\\ 
\midrule
Reward & GAIL& \makecell{MMP, LEARCH, \\ MaxEnt IOC, SQIL, \\\textbf{AdRIL}, A+N} \\
\midrule
On-$Q$  &  \makecell{DAgger, GPS \\ \textbf{iFAIL}, \textbf{DAeQuIL}}  & \xmark\\
\bottomrule
\end{tabular}
\end{sc}
\end{small}
\end{center}
\vskip -0.1in
\caption{An taxonomy of moment matching algorithms. \textbf{Bold} text indicates algorithms that are IPM-based. \label{onetabletorulethemall} }
\end{table}

\subsection{The Hidden Cost of Reward Moment Matching}
At first glance, the reward moment matching bound might seem to good to be true -- reward matching algorithms don't require a queryable expert like on-$Q$ approaches yet their performance bound seems to be tighter. This better performance characteristic is a product of a potentially \textit{exponentially} harder optimization problem for the learner. Consider the following tree-structured MDP with $|\mathcal{A}|$ actions at each step, each of which lead to a distinct state.
\begin{figure}[ht]
    \centering
    \begin{tikzpicture}[scale=1, transform shape]
    \node (b) [draw, very thick, circle, color=black] at (1.5, 1) {$s_0$};
    \node (c)  [circle]  at (1.5, 0) {$\ldots$}; 
    \node (d) [draw, very thick, circle, color=error]  at (0., -0) {$s_a$};
    \node (e) [draw, very thick, circle, color=expert]  at (3, -0) {$s_1$};
    
    \node (f) [draw, very thick, circle, color=error]  at (-1.5, -1) {$s_b$};
    \node (g)  [circle]  at (-0.75, -1) {$\ldots$}; 
    \node (h) [draw, very thick, circle, color=error]  at (0.0, -1) {$s_c$};
    \node (i) [draw, very thick, circle, color=error]  at (0.75, -1) {$s_d$};
    \node (j) [circle]  at (1.5, -1) {$\ldots$};
    \node (k) [draw, very thick, circle, color=error]  at (2.25, -1) {$s_e$};
    \node (l) [draw, very thick, circle, color=error]  at (3, -1) {$s_{f}$};
    \node (m) [circle]  at (3.75, -1) {$\ldots$};
    \node (n) [draw, very thick, circle, color=expert]  at (4.5, -1) {$s_{2}$};
    \node (o) [circle]  at (1.5, -1.5) {$\ldots$};
    
    \draw [->, very thick, color=error] (d) to (g);
    \draw [->, very thick, color=error] (d) to (f);
    \draw [->, very thick, color=error] (d) to (h);
    
    \draw [->, very thick, color=error] (c) to (i);
    \draw [->, very thick, color=error] (c) to (j);
    \draw [->, very thick, color=error] (c) to (k);
    
    \draw [->, very thick, color=error] (e) to (l);
    \draw [->, very thick, color=error] (e) to (m);
    \draw [->, very thick, color=expert] (e) to (n);
    \path[->, color=expert] (e) to node[midway, right] {$a_{|\mathcal{A}|}$} (n);

    \path[->, color=expert] (b) to node[midway, right] {$a_{|\mathcal{A}|}$} (e);
    \draw [->, very thick, color=expert] (b) to (e);
    
    \draw [->, very thick, color=error] (b) to (c);
    \draw [->, very thick, color=error] (b) to (d);
    \end{tikzpicture}
    \caption{In \textsc{Tree}, the expert always takes the right-most action from the current node. The exponential number of $T$-length trajectories in this problem make it a challenge for reward moment matching approaches.}
\end{figure}
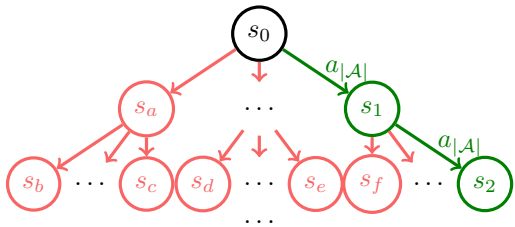
Consider an expert that takes $a_{|\mathcal{A}|}$ at each timestep. Solving the reward matching problem requires the learner to simultaneously optimize over all $T$ timesteps of the problem while considering the effect of past actions on future states. If we set $\Pi$ to be the class of deterministic policies, this is equivalent to optimizing over the set of all length $T$ trajectories, of which there are $O(|\mathcal{A}|^T)$ in tree-structured problems. In contrast, for off-$Q$ approaches, we attempt to match expert moments on a fixed expert state distribution. Similarly, we can optimize over a fixed history of past learner state distributions under a weak realizability assumption in the on-$Q$ setting. In the preceding example, the $Q$-matching settings are like being handed a node at each level of the tree and being asked to choose between the $|\mathcal{A}|$ edges available, leading to a total of $O(|\mathcal{A}|T)$ options to choose between. As we saw in Sec. \ref{sec:equib}, the policy player sometimes needs to compute a best response over the entire set of choices it has available, which means these search space sizes directly affect the per-iteration complexity of the moment-matching algorithms. Concisely, the price we pay for solving an easier optimization problem is looser policy performance bounds. 

\subsection{Takeaways}
In this work, we tease apart the differences in requirements and performance guarantees that come from matching reward, on-$Q$, and off-$Q$  adversarially chosen moments. Reward moment matching has strong guarantees but requires access to an accurate simulator or the real world. Off-$Q$ moment matching can be done purely on collected data but incurs an $O(\epsilon T^2)$ imitation gap. 

We formalize a notion of \textit{recoverability} that is both necessary and sufficient to understand recovering from errors in imitation learning. If a problem (due to expert or the MDP itself) is $O(T)$-recoverable, there exist problems where no algorithm can escape an $O(T^2)$ compounding of errors; if it is $O(1)$-recoverable, we find on policy algorithms prevent compounding. Together, these constitute a cohesive picture of moment matching in imitation learning.

We derive idealized no-regret procedures and practical IPM-based algorithms that are conceptually elegant and correct for difficulties encountered by prior methods. While behavioral cloning equally weights action-conditional errors, \texttt{AdVIL} can prevent headaches with value moment-based weighting. \texttt{AdRIL} is simple to implement, does not require training a GAN, and enjoys strong performance both in theory and practice. \texttt{DAeQuIL}'s moment-based losses are able to help relieve hiccups from focusing on action-conditionals that can stymie DAgger.


\section*{Acknowledgments}
We thank Siddharth Reddy for helpful discussions. We also thank Allie Del Giorno, Anirudh Vemula, and the members of the Social AI Group for their comments on drafts of this work. ZSW was supported in part by the NSF FAI Award \#1939606, a Google Faculty Research Award, a J.P. Morgan Faculty Award, a Facebook Research Award, and a Mozilla Research Grant. Lastly, GS would like to thank John Steinbeck for inspiring the title of this work:
\vspace{-1em}
\begin{quote}
\textit{``As happens sometimes, a moment settled and hovered and remained for much more than a moment."}
-- John Steinbeck, Of Mice and Men.
\end{quote}

\clearpage
\bibliographystyle{plainnat}
\bibliography{main}

\clearpage
\icmltitle{Supplementary Material}
\appendix

\section{Proofs}
\label{sec:proofs}
\subsection{Proof of Lemma \ref{thm:nonreclb}}
\begin{pf}
Consider the following MDP:
\begin{figure}[H]
    \centering
    \begin{tikzpicture}[scale=1, transform shape]
    \node (b) [draw, very thick, circle, color=expert] at (1.5, 1) {$s_1$};
    \node (c) [draw, very thick, circle, color=error]  at (1.5, -1) {$s_2$};
    \path[->, color=error] (b) to node[midway, right] {$a_2$} (c);
    \draw [->, very thick, color=error] (b) to (c);
    \node [circle, minimum size=0.5cm](d) at ([{shift=(90:0.4)}]b){};
    \coordinate (e) at (intersection 2 of b and d);
    \coordinate (f) at (intersection 1 of b and d);
    \tikzAngleOfLine(d)(f){\AngleStart}
    \tikzAngleOfLine(d)(e){\AngleEnd}
    \draw[very thick,->, color=expert]%
    let \p1 = ($ (d) - (f) $), \n2 = {veclen(\x1,\y1)}
    in
        (d) ++(90:0.4) node{$a_1$}
        (f) arc (\AngleStart-360:\AngleEnd:\n2);
        
    \node [circle, minimum size=0.5cm](g) at ([{shift=(270:0.4)}]c){};
    \coordinate (h) at (intersection 2 of c and g);
    \coordinate (i) at (intersection 1 of c and g);
    \tikzAngleOfLine(g)(i){\AngleStart}
    \tikzAngleOfLine(g)(h){\AngleEnd}
    \draw[very thick,->,]%
    let \p1 = ($ (g) - (i) $), \n2 = {veclen(\x1,\y1)}
    in
        (g) ++(270:0.5) node{$a_1$}
        (i) arc (\AngleStart-360:\AngleEnd:\n2);
    \end{tikzpicture}
    \caption{In \textsc{Unicycle}, the goal is to stay in $s_1$ forever.}
\end{figure}
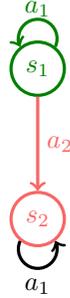
As illustrated above, $c(s, a) = \mathds{1}_{s_2}$ is the cost function for this MDP. Let the expert perfectly optimize this function by always taking $a_1$ in $s_1$. Thus, we are in the $O(T)$-recoverable setting. Then, for any $\epsilon > 0$, if the learner takes $a_2$ in $s_1$ with probability $\epsilon$, $J(\pi_E) - J(\pi) = \sum_{t=1}^T \epsilon (1 - \epsilon)^{t-1}(T - t) = \Omega(\epsilon T^2)$. There is only one action in $s_2$ so it is not possible to have a nonzero classification error in this state.
\end{pf}
\label{proof:nonreclb}
\subsection{Proof of Entropy Regularization Lemma}
\begin{lemma}
\label{lem:entreg} {\textbf{Entropy Regularization Lemma:}}
By optimizing $U_j(\pi, f) - \alpha H(\pi)$ to a $\delta$-approximate equilibrium, one recovers at worst a $Q_M\sqrt{\frac{2 \delta}{\alpha}} + \alpha T(\ln{|\mathcal{A}|}+\ln{|\mathcal{S}|)}$ equilibrium strategy for the policy player on the original game.
\end{lemma}
\label{proof:entreg}
\begin{pf}
We optimize in the $P(\mathbf{A}^T||\mathbf{S}^T)$ policy representation where strong duality holds and  define the following:
\begin{flalign*}
  \begin{aligned}
 \pi^R &= \argmin_{\pi \in \Pi} (\max_{f \in \mathcal{F}} U_j(\pi, f) - \alpha H(\pi)) \\
  \end{aligned}
\end{flalign*}
First, we derive a bound on the distance between $\hat{\pi}$ and $\pi^R$. We define $M$ as follows:
\begin{flalign*}
  \begin{aligned}
 M(\pi) &= \max_{f \in \mathcal{F}} U_j(\pi, f) - \alpha H(\pi) + \alpha T (\ln{|\mathcal{A}|} + \ln{|\mathcal{S}|)}\\
  \end{aligned}
\end{flalign*}
$M$ is an $\alpha$-strongly convex function with respect to $||\cdot||_1$ because $U$ is a max of linear functions, $-H$ is 1-strongly convex, and the third term is a constant. This tells us that:
\begin{flalign*}
  \begin{aligned}
 M(\pi^R) - M(\hat{\pi}) \leq \nabla M(\pi^R)^T (\pi^R - \hat{\pi}) - \frac{\alpha}{2}||\pi^R - \hat{\pi}||_1^2
  \end{aligned}
\end{flalign*}
We note that because $\pi^R$ minimizes $M$, the first term on the RHS is negative, allowing us to simplify this expression to:
\begin{flalign*}
  \begin{aligned}
 \frac{\alpha}{2}||\pi^R - \hat{\pi}||_1^2 \leq  M(\hat{\pi}) - M(\pi^R)
  \end{aligned}
\end{flalign*}
We now upper bound the RHS of this expression via the following series of substitutions:
\begin{flalign*}
  \begin{aligned}
 M(\hat{\pi}) &= \max_{f \in \mathcal{F}} U_j(\hat{\pi}, f) - \alpha H(\hat{\pi}) + \alpha T (\ln{|\mathcal{A}|} + \ln{|\mathcal{S}|)} \\
 & \leq U_j(\hat{\pi}, \hat{f}) - \alpha H(\hat{\pi}) + \alpha T (\ln{|\mathcal{A}|} + \ln{|\mathcal{S}|)} + \delta \\
 & \leq U_j(\pi^R, \hat{f}) - \alpha H(\pi^R) + \alpha T (\ln{|\mathcal{A}|} + \ln{|\mathcal{S}|)} + \delta \\
 & \leq \max_{f \in \mathcal{F}} U_j(\pi^R, f) - \alpha H(\pi^R) + \alpha T (\ln{|\mathcal{A}|} + \ln{|\mathcal{S}|)} + \delta \\
 & = M(\pi^R) + \delta
  \end{aligned}&
\end{flalign*}
Rearranging terms to get the desired bound on strategy distance:
\begin{flalign*}
  \begin{aligned}
  & M(\hat{\pi}) - M(\pi^R) \leq \delta \\
  & \Rightarrow ||\pi^R - \hat{\pi}||_1^2 \leq \frac{2 \delta}{\alpha} \\
  & \Rightarrow ||\pi^R - \hat{\pi}||_1 \leq \sqrt{\frac{2 \delta}{\alpha}}
  \end{aligned}&
\end{flalign*}
Next, we prove that $\pi^R$ is a $\alpha T (\ln{|\mathcal{A}|} + \ln{|\mathcal{S}|)}$-approximate equilibrium strategy for the original, unregularized game. We note that $H(\pi) \in [0, T(\ln{|\mathcal{A}|}+\ln{|\mathcal{S}|)}]$ and then proceed as follows:
\begin{flalign*}
  \begin{aligned}
  \max_{f \in \mathcal{F}} U_j(\pi^R, f) &= M(\pi^R) + \alpha H(\pi^R) - \alpha T (\ln{|\mathcal{A}|} + \ln{|\mathcal{S}|)} \\
  & \leq M(\pi^R) \\
  & \leq \alpha T(\ln{|\mathcal{A}|}+\ln{|\mathcal{S}|)}
  \end{aligned}&
\end{flalign*}
The last line comes from the fact that playing the optimal strategy in the original game on the regularized game could at worst lead to a payoff of $\alpha T(\ln{|\mathcal{A}|}+\ln{|\mathcal{S}|)}$. Therefore, the value of the regularized game can at most be this quantity. Recalling that the value of the original game is 0 and rearranging terms, we get:
\begin{flalign*}
  \begin{aligned}
  \max_{f \in \mathcal{F}} U_j(\pi^R, f) -\alpha T(\ln{|\mathcal{A}|}+\ln{|\mathcal{S}|)} \leq 0 = \max_{f \in \mathcal{F}} \min_{\pi \in \Pi} U_j(\pi, f)
  \end{aligned}&
\end{flalign*}
Thus by definition, $\pi^R$ must be half of an $\alpha T(\ln{|\mathcal{A}|}+\ln{|\mathcal{S}|)}$-approximate equilibrium strategy pair. 

Next, let $Q_M$ denote the absolute difference between the minimum and maximum $Q$-value. For a fixed $f$, the maximum amount the policy player could gain from switching to policies within an $L_1$ ball of radius $r$ centered at the original policy is $r Q_M$ by the bilinearity of the game and Hölder's inequality. Because the supremum over $k$-Lipschitz functions is known to be $k$-Lipschitz, this implies the same is true for the payoff against the best response $f$. To complete the proof, we can set $r = \sqrt{\frac{2 \delta}{\alpha}}$ and combine this with the fact that $\pi^R$ achieves in the worst case a payoff of $\alpha T(\ln{|\mathcal{A}|}+\ln{|\mathcal{S}|})$ to prove that $\hat{\pi}$ can at most achieve a payoff of $Q_M\sqrt{\frac{2 \delta}{\alpha}} + \alpha T(\ln{|\mathcal{A}|}+\ln{|\mathcal{S}|)}$ on the original game, which establishes $\hat{\pi}$ as a $(Q_M\sqrt{\frac{2 \delta}{\alpha}} + \alpha T(\ln{|\mathcal{A}|}+\ln{|\mathcal{S}|)})$-approximate equilibrium solution.

\end{pf}
\subsection{Proof of Theorem \ref{thm:master}}
\label{proof:master}
We proceed in cases.
\begin{pf} We first consider the \textbf{primal case}. Our goal is to compute a policy $\hat{\pi}$ such that:
$$
\max_{f \in \mathcal{F}} U_j(\hat{\pi}, f) \leq \delta
$$
We prove that such a policy can be found efficiently by executing the following procedure for a polynomially large number of iterations:
\begin{enumerate}
\setlength\itemsep{0em}
    \item For $t = 1 \dots N$, do:
    {\setlength\itemindent{25pt}
    \item No-regret algorithm computes $\pi^t$.
    \item Set $f^t$ to the best response to $\pi^t$.
    }
    \item Return $\hat{\pi} = \pi^{t^*}$, $t^* = \argmin_{t} U_j(\pi^t, f^t)$.
\end{enumerate}

Recall that via our no-regret assumption we know that
$$
\frac{1}{N} \sum_t^N U_j(\pi^t, f^t) - \frac{1}{N} \min_{\pi \in \Pi} \sum_t^N U_j(\pi, f^t) \leq \frac{\beta_{\Pi}(N)}{N} \leq \delta
$$
for some $N$ that is $\text{poly}(\frac{1}{\delta})$. We can rearrange terms and use the fact that $\pi_E \in \Pi$ to upper bound the average payoff:
$$
\frac{1}{N} \sum_t^N U_j(\pi^t, f^t) \leq \delta + \frac{1}{N} \min_{\pi \in \Pi} \sum_t^N U_j(\pi, f^t) \leq \delta
$$
Using the property that there must be at least one element in an average that is at most the value of the average:
$$
\min_t U_j(\pi^t, f^t) \leq \frac{1}{N} \sum_t^N U_j(\pi^t, f^t) \leq \delta
$$
To complete the proof, we recall that $f^t$ is chosen as the best response to $\pi^t$, giving us that:
$$
\min_t \max_{f \in \mathcal{F}} U_j(\pi^t, f) \leq \delta
$$
In words, this means that by setting $\hat{\pi}$ to the policy with the lowest loss out of the $N$ computed, we are able to efficiently (within $\text{poly}(\frac{1}{\delta})$ iterations) find a $\delta$-approximate equilibrium strategy for the policy player. Note that this result holds without assuming a finite $\mathcal{S}$ and $\mathcal{A}$ and does not require regularization of the policy. However, it requires us to have a no-regret algorithm over $\Pi$ which can be a challenge for the reward moment-matching game.

We now consider the \textbf{dual case}. As before, we wish to find a policy $\hat{\pi}$ such that:
$$
\max_{f \in \mathcal{F}} U_j(\hat{\pi}, f) \leq \delta
$$
We run the following procedure on $U_j(\pi, f) - \alpha H(\pi)$:
\begin{enumerate}
\setlength\itemsep{0em}
    \item For $t = 1 \dots N$, do:
    {\setlength\itemindent{25pt}
    \item No-regret algorithm computes $f^t$.
    \item Set $\pi^t$ to the best response to $f^t$.
    }
    \item Return $\hat{\pi} = \argmin_{\pi \in \Pi} U_j(\pi, \overbar{f}) - \alpha H(\pi)$.
\end{enumerate}
By the classic result of \cite{freund1997decision}, we know that the average of the $N$ iterates produced by the above procedure (which we denote $\overbar{f}$ and $\overbar{\pi}$) is a $\delta'$-approximate equilibrium strategy for some $N$ that is $\text{poly}(\frac{1}{\delta'})$. Applying our Entropy Regularization Lemma, we can upper bound the payoff of $\overbar{\pi}$ on the original game:
$$
\sup_{f \in \mathcal{F}} U_j(\overbar{\pi}, f) \leq Q_M\sqrt{\frac{2 \delta'}{\alpha}} + \alpha T(\ln{|\mathcal{A}|}+\ln{|\mathcal{S}|)}
$$
We now proceed similarly to our proof of the Entropy Regularization Lemma by first bounding the distance between $\overbar{\pi}$ and $\hat{\pi}$ and the appealing to the $Q_m$-Lipschitzness of $U_j$. Let $l(\pi) = U_j(\pi, \overbar{f}) - \alpha H(\pi)$. Then, while keeping the fact that $l$ is $\alpha$-strongly convex in mind:
\begin{flalign*}
  \begin{aligned}
   & l(\hat{\pi}) - l(\overbar{\pi}) \leq \nabla l(\hat{\pi})^T (\hat{\pi} - \overbar{\pi}) - \frac{\alpha}{2}||\overbar{\pi} - \hat{\pi}||_1^2 \\
   & \Rightarrow \frac{\alpha}{2}||\overbar{\pi} - \hat{\pi}||_1^2 \leq l(\overbar{\pi}) - l(\hat{\pi}) + \nabla l(\hat{\pi})^T (\hat{\pi} - \overbar{\pi}) \\
   & \Rightarrow \frac{\alpha}{2}||\overbar{\pi} - \hat{\pi}||_1^2 \leq l(\overbar{\pi}) - l(\hat{\pi}) \\ & \Rightarrow \frac{\alpha}{2}||\overbar{\pi} - \hat{\pi}||_1^2 \leq \delta'
   \\ & \Rightarrow ||\overbar{\pi} - \hat{\pi}||_1 \leq \sqrt{\frac{2 \delta'}{\alpha}}
  \end{aligned}&
\end{flalign*}
As before, the second to last step follows from the definition of a $\delta'$-approximate equilibrium. Now, by the bilinearity of the game, Hölder's inequality, and the fact that supremum over $k$-Lipschitz functions is known to be $k$-Lipschitz, we can state that:
$$
\sup_{f \in \mathcal{F}} U_j(\hat{\pi}, f) \leq 2 Q_M\sqrt{\frac{2 \delta'}{\alpha}} + \alpha T(\ln{|\mathcal{A}|}+\ln{|\mathcal{S}|)}
$$
To ensure that the LHS of this expression is upper bounded by $\delta$, it is sufficient to set $\alpha = \frac{\delta}{2 T(\ln{\mathcal{|A|}} + \ln{\mathcal{|S|}})}$ and $\delta' = \frac{\delta^2 \alpha}{32 Q_M^2}$. Plugging in these terms, we arrive at:
$$
\sup_{f \in \mathcal{F}} U_j(\hat{\pi}, f) \leq \frac{\delta}{2} + \frac{\delta}{2} \leq \delta
$$
We note that in practice, $\alpha$ is rather sensitive hyperparameter of maximum entropy reinforcement learning algorithms \cite{haarnoja2018soft} and hope that the above expression might provide some rough guidance for how to set $\alpha$. To complete the proof, note that $N$ is $\text{poly}(\frac{1}{\delta'})$ and $\frac{1}{\delta'} = \frac{64 Q_M^2 T(\ln{|\mathcal{A}|}+\ln{|\mathcal{S}|)}}{\delta^3}$. Thus, $N$ is $\text{poly}(\frac{1}{\delta}, T, \ln{|\mathcal{A}|}, \ln{|\mathcal{S}|)})$.

\end{pf}

\section{Algorithm Derivations}
\subsection{\texttt{AdVIL} Derivation}
\label{advil_derivation}
We begin by performing the following substitution: $f = v - \mathcal{B}^{\pi}v$, where $$\mathcal{B}^{\pi} v = \mathop{{}\mathbb{E}}_{\substack{s_{t+1} \sim \mathcal{T}(\cdot | s_t, a_t), \\ a_{t+1} \sim \pi(s_{t+1})}}[v]$$ is the expected Bellman operator under the learner's current policy. Our objective \eqref{eq:ipm} then becomes:
\begin{flalign*}
  \begin{aligned}
    \sup_{v \in \mathcal{F}} \sum_{t = 1}^T &  \mathop{{}\mathbb{E}}_{\tau \sim \pi}[v(s_t, a_t) - \mathcal{B}^{\pi}v(s_t, a_t)]\\
    & - \mathop{{}\mathbb{E}}_{\tau \sim \pi_E}[v(s_t, a_t) - \mathcal{B}^{\pi}v(s_t, a_t)]
  \end{aligned}
\end{flalign*}

This expression telescopes over time, simplifying to:
$$
\sup_{v \in \mathcal{F}} \mathop{{}\mathbb{E}}_{\tau \sim \pi}[v(s_0, a_0)] - \sum_{t = 1}^T  \mathop{{}\mathbb{E}}_{\tau \sim \pi_E}[v(s_t, a_t) - \mathcal{B}^{\pi}v(s_t, a_t)]
$$
\\
We approximate $\mathcal{B}^{\pi} v$ via a single-sample estimate from the respective expert trajectory, yielding the following off-policy expression:
$$
\sup_{v \in \mathcal{F}} \mathop{{}\mathbb{E}}_{\tau \sim \pi}[v(s_0, a_0)] - \sum_{t = 1}^T  \mathop{{}\mathbb{E}}_{\tau \sim \pi_E}[v(s_t, a_t) - \mathop{{}\mathbb{E}}_{a \sim \pi(s_{t+1})}[v(s_{t+1}, a)]]
$$
This resembles the form of the objective in ValueDICE \cite{kostrikov2019imitation} but without requiring us to take the expectation of the exponentiated discriminator. We can further simplify this objective by noticing that trajectories generated by $\pi_E$ and $\pi$ have the same starting state distribution:
\begin{equation}
\sup_{v \in \mathcal{F}} 
\mathop{{}\mathbb{E}}_{\tau \sim \pi_E}[   
\sum_{t = 1}^T  \mathop{{}\mathbb{E}}_{a \sim \pi(s_{t})}[v(s_{t}, a)]
- v(s_t, a_t)]
\end{equation}

We also note that this \texttt{AdVIL} objective can be derived straightforwardly via the Performance Difference Lemma.

\subsection{\texttt{AdRIL} Derivation}
\label{adril_derivation}
 Let $\mathcal{F}$ be a RKHS be equipped with kernel $K: (\mathcal{S} \times \mathcal{A}) \times (\mathcal{S} \times \mathcal{A}) \rightarrow \mathbb{R}$. On iteration $k$ of the algorithm, consider a purely cosmetic variation of our IPM-based objective \eqref{eq:ipm}:
$$
\sup_{c \in \mathcal{F}} \sum_{t = 1}^T (\mathop{{}\mathbb{E}}_{\tau \sim \pi_k}[c(s_t, a_t)] - \mathop{{}\mathbb{E}}_{\tau \sim \pi_E}[c(s_t, a_t)]) = \sup_{c \in \mathcal{F}} L_k(c)
$$
We evaluate the first expectation by collecting on-policy rollouts into a dataset $\mathcal{D}_k$ and the second by sampling from a fixed set of expert demonstrations $\mathcal{D}_E$. Assume that $|\mathcal{D}_k|$ is constant across iterations. Let $\mathcal{E}$ be the evaluation functional. Then, taking the functional gradient:
\begin{flalign*}
  \begin{aligned}
\nabla_c L_k(c) &=  \sum_{t = 1}^T  \frac{1}{|\mathcal{D}_k|}\sum_{\tau}^{\mathcal{D}_k}\nabla_c\mathcal{E}[c; (s_t, a_t)] - \frac{1}{|\mathcal{D}_E|} \sum_{\tau}^{\mathcal{D}_E}\nabla_c\mathcal{E}[c; (s_t, a_t)] \\
    & = \sum_{t = 1}^T  \frac{1}{|\mathcal{D}_k|}\sum_{\tau}^{\mathcal{D}_k}K([s_t, a_t], \cdot) -  \frac{1}{|\mathcal{D}_E|}\sum_{\tau}^{\mathcal{D}_E}K([s_t, a_t], \cdot)
  \end{aligned}
\end{flalign*}
where $K$ could be an state-action indicator ($\mathds{1}_{s, a}$) in discrete spaces and relaxed to a Gaussian in continuous spaces. Let $\overbar{\mathcal{D}_k} = \bigcup_{i=0}^k \mathcal{D}_i$ be the aggregation of all previous $\mathcal{D}_i$. Averaging functional gradients over iterations of the algorithm (which, other than a scale factor that does not affect the optimal policy, is equivalent to having a constant learning rate of $1$), we get the cost function our policy tries to minimize:
\begin{equation}
  \begin{aligned}
C(\pi_k) &=  \sum_{i = 0}^k \nabla_c L_i(c) \\
    & = \sum_{t = 1}^T  \frac{1}{|\overbar{\mathcal{D}_k}|}\sum_{\tau}^{\overbar{\mathcal{D}_k}}K([s_t, a_t], \cdot) - \frac{1}{|\mathcal{D}_E|} \sum_{\tau}^{\mathcal{D}_E}K([s_t, a_t], \cdot)\label{eq:adril}
  \end{aligned}
\end{equation}

\subsection{\texttt{DAeQuIL} Derivations}
\label{daequil_derivation}
Let $d_{\pi}$ denote the state-action visitation distribution of $\pi$. Then, \texttt{DAeQuIL} can be seen as Follow The Regularized Leader on the following sequence of losses:
\begin{enumerate}
    \item $f_i = \argmax_{f \in \mathcal{F}} \mathbb{E}_{s, a \sim d_{\pi_i}}[f(s, a) - f(s, \pi_E(s))]$
    \item $l_i(\pi) = \mathbb{E}_{s \sim d_{\pi_i}}[f_i(s, \pi(s)) - f_i(s, \pi_E(s))]$
\end{enumerate}
Solving the on-$Q$ game proper would instead require $l'_i(\pi) = \mathbb{E}_{s \sim d_{\pi}}[f_i(s, \pi(s)) - f_i(s, \pi_E(s))]$ -- for the state distribution to depend on the policy that is passed to the loss. While this would allow our previous no-regret analysis to apply as written, we would need to re-sample trajectories after every gradient step, a burden we'd like to avoid.

Let us consider the no-regret guarantee we get from the \texttt{DAeQuIL} losses:
$$
\frac{1}{N} \sum_t^N l_t(\pi^t) - \frac{1}{N} \min_{\pi \in \Pi} \sum_t^N l_t(\pi) \leq \frac{\beta_{\Pi}(N)}{N} \leq \delta
$$

Notice that $l_t(\pi^t) = \max_{f \in \mathcal{F}} U_3(\pi^t, f)$, the exact quantity we'd like to bound. The tricky part comes from the second term in the regret -- under realizability, ($\pi_E \in \Pi$), this term is 0 and \texttt{DAeQuIL} directly finds a $\delta$-approximate equilibrium for the on-$Q$ game. Otherwise, we require the following weak notion of realizability to maintain the on-$Q$ moment matching bounds: $\exists \pi' \in \Pi$ s.t. 
$$\max_{d_{\pi} \in d_{\Pi}} \max_{f \in \mathcal{F}} \mathbb{E}_{s \sim d_{\pi}}[f(s, \pi'(a)) - f(s, \pi_E(a))] \leq O(\epsilon)$$
In words, this is saying that there exists a policy $\pi'$ that can match expert moments up to $\epsilon$ on any state visitation distribution generated by a policy in $\Pi$. If we instead solved the on-$Q$ game directly by using $l'_i(\pi)$, we would instead need the condition: $\exists \pi' \in \Pi$ s.t.
$$\max_{f \in \mathcal{F}} \mathbb{E}_{s \sim d_{\pi'}}[f(s, \pi'(a)) - f(s, \pi_E(a))] \leq O(\epsilon)$$
This weaker condition is concomitant with a much more computationally expensive optimization procedure.

\section{Experimental Setup}
\label{params}
\subsection{Expert}
We use the Stable Baselines 3 \cite{stable-baselines3} implementation of PPO \cite{schulman2017proximal} and SAC \cite{haarnoja2018soft} to train experts for each environment, mostly using the already tuned hyperparameters from \cite{rl-zoo3}. Specifically, we use the modifications in Tables 4 and 5 to the Stable Baselines Defaults.
\begin{table}[h]
\begin{center}
\begin{small}
\begin{sc}
\setlength{\tabcolsep}{2pt}
\begin{tabular}{lccccr}
\toprule
 Parameter & Value \\
\midrule
 Buffer Size & 300000 \\
 Batch Size & 256 \\
 $\gamma$ & 0.98 \\
 $\tau$ & 0.02 \\
 Training Freq. & 64 \\
 Gradient Steps & 64 \\
 Learning Rate & 7.3e-4 \\
 Policy Architecture & 256 x 2 \\
 State-Dependent Exploration & True \\
 Training Timesteps & 1e6 \\
\bottomrule
\end{tabular}
\end{sc}
\end{small}
\end{center}
\caption{Expert hyperparameters for HalfCheetah Bullet Task.}
\end{table}
\begin{table}[h]
\begin{center}
\begin{small}
\begin{sc}
\setlength{\tabcolsep}{2pt}
\begin{tabular}{lccccr}
\toprule
 Parameter & Value \\
\midrule
 Buffer Size & 300000 \\
 Batch Size & 256 \\
 $\gamma$ & 0.98 \\
 $\tau$ & 0.02 \\
 Training Freq. & 64 \\
 Gradient Steps & 64 \\
 Learning Rate & 7.3e-4 \\
 Policy Architecture & 256 x 2 \\
 State-Dependent Exploration & True \\
 Training Timesteps & 1e6 \\
\bottomrule
\end{tabular}
\end{sc}
\end{small}
\end{center}
\caption{Expert hyperparameters for Ant Bullet Task.}
\end{table}
\begin{table}
\begin{center}
\begin{small}
\begin{sc}
\setlength{\tabcolsep}{2pt} 
\begin{tabular}{lccccr}
\toprule
 Env. & Expert & BC Performance\\
\midrule
 HalfCheetah & 2154 & 2083 \\
 Ant & 2585 & 2526 \\
\bottomrule
\end{tabular}
\end{sc}
\end{small}
\end{center}
\caption{With enough data (25 trajectories) and 100k steps of gradient descent, behavioral cloning is able to solve all tasks considered, replicating the results of \cite{ALICE}. However, other approaches are able to perform better when there is less data available.}
\end{table}

\subsection{Baselines}
For all learning algorithms, we perform 5 runs and use a common architecture of 256 x 2 with ReLU activations. For each datapoint, we average the cumulative reward of 10 trajectories. For offline algorithms, we train on $\{5, 10, 15, 20, 25\}$ expert trajectories with a maximum of 500k iterations of the optimization procedure. For online algorithms, we train on a fixed number of trajectories (5 for HalfCheetah and 20 for Ant) for 500k environment steps. For GAIL \cite{ho2016generative} and behavioral cloning \cite{pomerleau1989alvinn}, we use the implementation produced by \cite{wang2020imitation}. We use the changes from the default values in Tables 6 and 7 for all tasks.
\begin{table}[h]
\begin{center}
\begin{small}
\begin{sc}
\setlength{\tabcolsep}{2pt}
\begin{tabular}{lccccr}
\toprule
 Parameter & Value \\
\midrule
 Entropy Weight & 0 \\
 L2 Weight & 0 \\
 Training Timesteps & 5e5 \\
\bottomrule
\end{tabular}
\end{sc}
\end{small}
\end{center}
\caption{Learner hyperparameters for Behavioral Cloning.}
\end{table}
\begin{table}[h]
\begin{center}
\begin{small}
\begin{sc}
\setlength{\tabcolsep}{2pt}
\begin{tabular}{lccccr}
\toprule
 Parameter & Value \\
\midrule
 Num Steps & 1024 \\
 Expert Batch Size & 32 \\
\bottomrule
\end{tabular}
\end{sc}
\end{small}
\end{center}
\caption{Learner hyperparameters for GAIL.}
\end{table}

For SQIL \cite{reddy2019sqil}, we build a custom implementation on top of Stable Baselines with feedback from the authors. As seen in Table 9, we use the similar parameters for SAC as we did for training the expert.

We modify the open-sourced code for ValueDICE \cite{kostrikov2019imitation} to be actually off-policy with feedback from the authors. The publicly available version of the ValueDICE code uses on-policy samples to compute a regularization term, even when it is turned off in the flags. We release our version.\footnote{\url{https://github.com/gkswamy98/valuedice}} We use the default hyperparameters for all experiments (and thus, train for 500k steps).

\subsection{Our Algorithms}
\begin{table}[ht]
\begin{center}
\begin{small}
\begin{sc}
\setlength{\tabcolsep}{2pt}
\begin{tabular}{lccccr}
\toprule
 Parameter & Value \\
\midrule
 $\gamma$ & 0.98 \\
 $\tau$ & 0.02 \\
 Training Freq. & 64 \\
 Gradient Steps & 64 \\
 Learning Rate & Linear Schedule of 7.3e-4 \\
\bottomrule
\end{tabular}
\end{sc}
\end{small}
\end{center}
\caption{Learner hyperparameters for SQIL.}
\end{table}

In this section, we use \textbf{bold text} to highlight sensitive hyperparameters. Similarly to SQIL, \texttt{AdRIL} is built on top of the Stable Baselines implementation of SAC. \texttt{AdVIL} is written in pure PyTorch. We use the same network architecture choices as for the baselines. For \texttt{AdRIL} we use the hyperparameters in Table 10 across all experiments.
\begin{table}[h]
\begin{center}
\begin{small}
\begin{sc}
\setlength{\tabcolsep}{2pt}
\begin{tabular}{lccccr}
\toprule
 Parameter & Value \\
\midrule
 $\gamma$ & 0.98 \\
 $\tau$ & 0.02 \\
 Training Freq. & 64 \\
 Gradient Steps & 64 \\
 Learning Rate & Linear Schedule of 7.3e-4 \\
 \textbf{ $f$ Update Freq.} & {1250} \\
\bottomrule
\end{tabular}
\end{sc}
\end{small}
\end{center}
\caption{Learner hyperparameters for \texttt{AdRIL}.}
\end{table}

We note that \texttt{AdRIL} requires careful tuning of \textbf{$f$ Update Freq.} for strong performance. To find the value specified, we ran trials with $\{1250, 2500, 5000, 12500, 25000, 50000\}$ and selected the one that achieved the most stable updates. In practice, we would recommend evaluating a trained policy on a validation set to set this parameter. We also note because SAC is an off-policy algorithm, we are free to initialize the learner by adding all expert samples to the replay buffer at the start, as is done for SQIL.

We change one parameter between environments for \texttt{AdRIL} -- for HalfCheetah, we perform standard sampling from the replay buffer while for Ant we sample an expert trajectory with $p=\frac{1}{2}$ and a learner trajectory otherwise, similar to SQIL. We find that for certain environments, this modification can somewhat increase the stability of updates while for other environments it can significantly hamper learner performance. We recommend trying both options if possible but defaulting to standard sampling.

For \texttt{AdVIL}, we use the hyperparameters in Table 11 across all tasks.
\begin{table}[h]
\begin{center}
\begin{small}
\begin{sc}
\setlength{\tabcolsep}{2pt}
\begin{tabular}{lccccr}
\toprule
 Parameter & Value \\
\midrule
 $\eta_{\pi}$ & 8e-6 \\
 $\eta_{f}$ & 8e-4 \\
 Batch Size & 1024 \\
 $f$ Gradient Target & 0.4 \\
 $f$ Gradient Penalty Weight & 10 \\
 $\pi$ Orthogonal Regularization & 1e-4 \\
 $\pi$ MSE Regularization Weight & 0.2 \\
 Normalize States with Expert Data & True \\
 Normalize Actions to [-1, 1] & True \\
 Gradient Norm Clipping & [-40, 40] \\
\bottomrule
\end{tabular}
\end{sc}
\end{small}
\end{center}
\caption{Learner hyperparameters for \texttt{AdVIL}.}
\end{table}
Emperically, small learning rates, large batch sizes, and regularization of both players are critical to stable convergence. We find that \texttt{AdVIL} converges significantly more quickly than ValueDICE, requiring only \textbf{50k steps for HalfCheetah and 100k Steps for Ant} instead of 500k steps for both tasks. However, we also find that running \texttt{AdVIL} for longer than these prescribed amounts can lead to a collapse of policy performance. Fortunately, this can easily be caught by watching for sudden and large fluctuations in policy loss after a long period of steady decreases. One can perform this early-stopping check without access to the environment.

\begin{figure}
    \centering
    \includegraphics[width=\columnwidth]{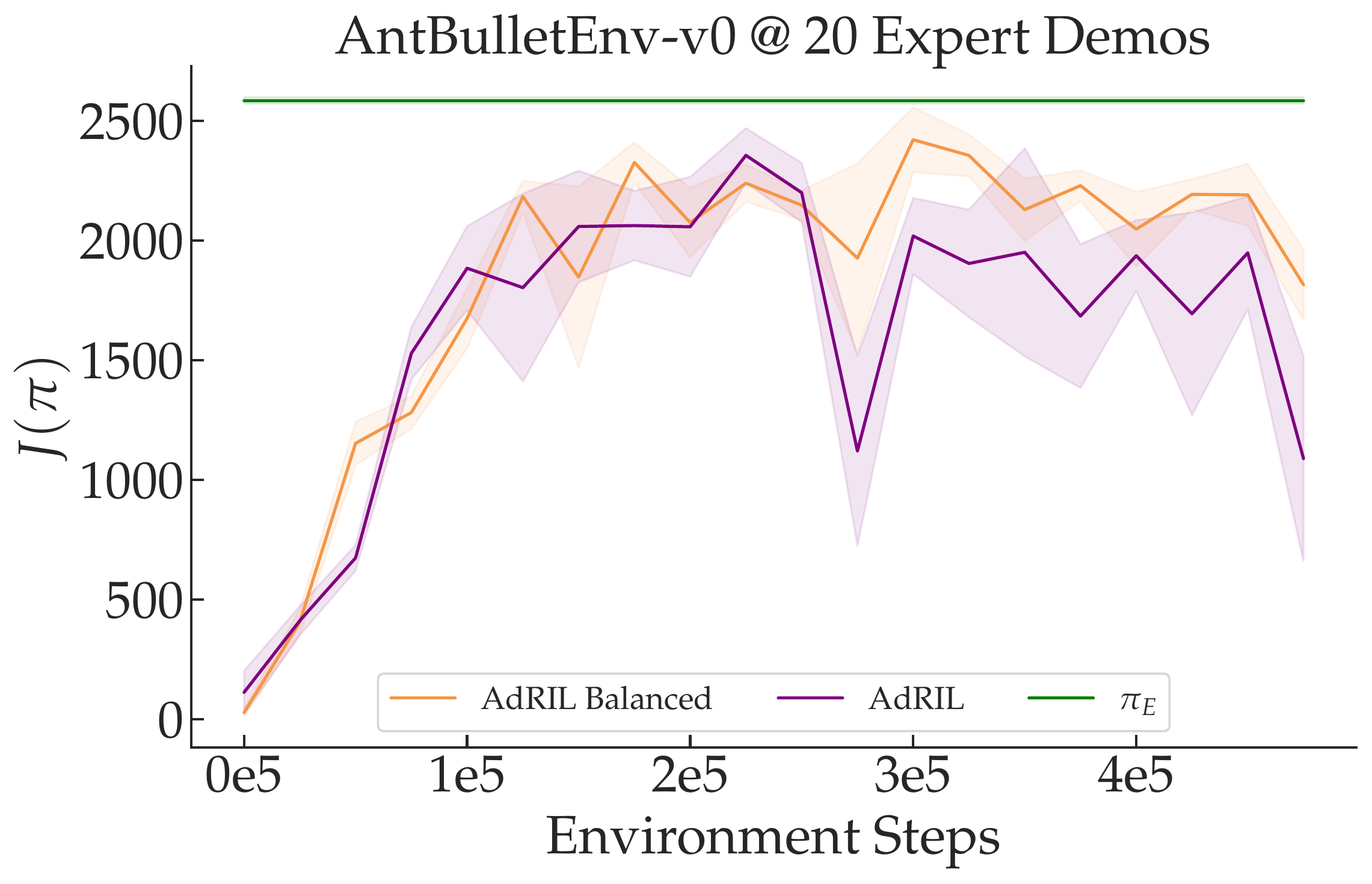}
    \caption{A comparison of balanced vs. unbalanced sampling for \texttt{AdRIL} on the Ant Environment. For certain tasks, balanced sampling can help with the stability of updates.}
\end{figure}

\section{On-$Q$ Experiments}
\label{sec:onq_exp}
\begin{figure}[ht]
    \centering
    \includegraphics[width=2\columnwidth/3]{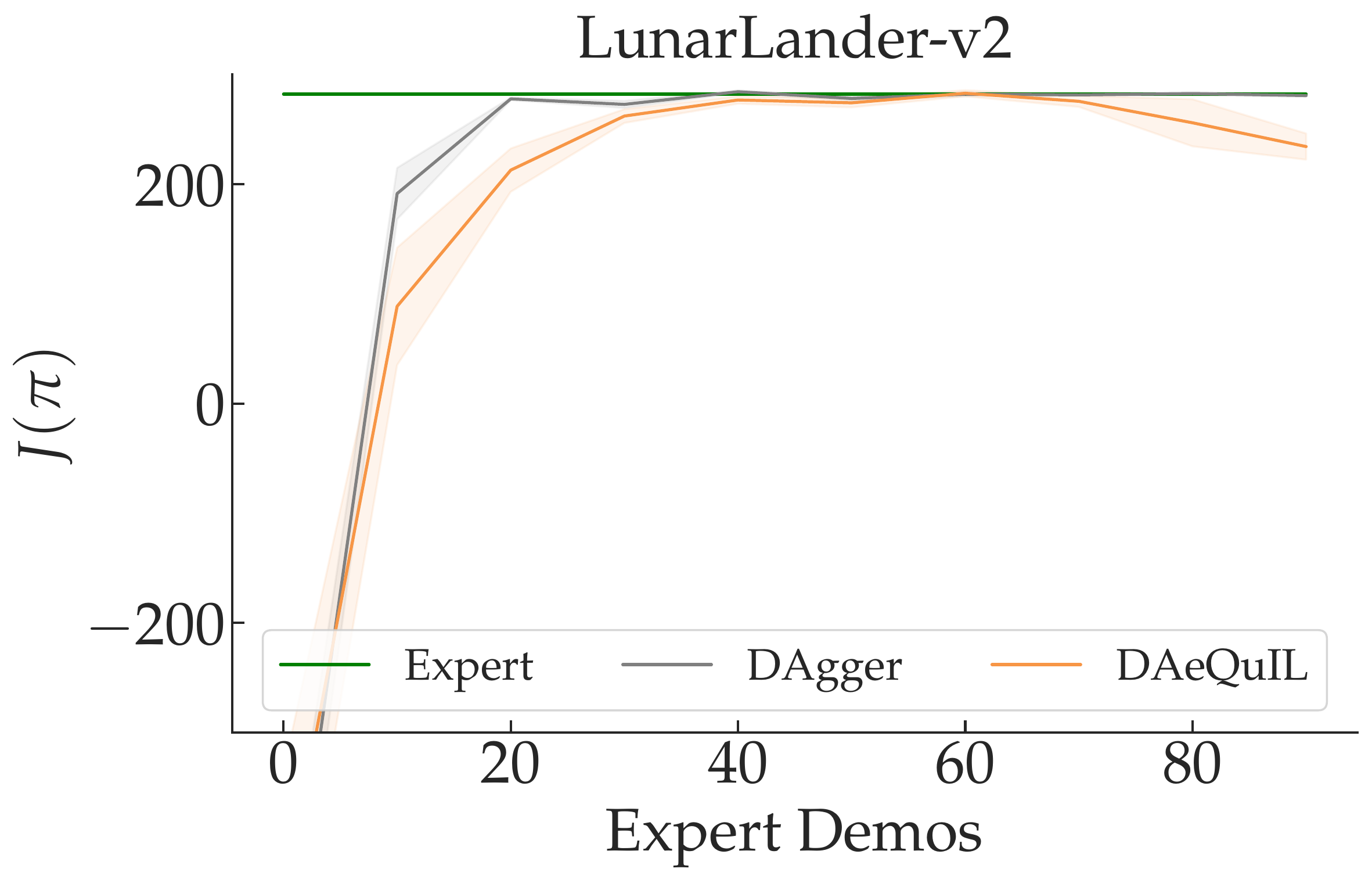}
    \caption{As behavioral cloning alone is able to nearly match the expert, DAgger and \texttt{DAeQuIL} perform around the same.}
    \label{fig:ll_iact}
\end{figure}

We perform two experiments to tease out when one should apply \texttt{DAeQuIL} over DAgger. We first present results on a rocket-landing task from OpenAI Gym where behavioral cloning by itself is able to nearly solve the task, as has been previously noted \cite{ALICE}. To make the task more challenging, we truncate the last two dimensions of the state for the policy class, which corresponds to masking the location of the legs of the lander. We use two-layer neural networks with 64 hidden units as all our function classes, perform the optimization steps via ADAM with learning rate $3e-4$, and sample 10 trajectories per update. Here, we see \texttt{DAeQuIL} do around as well as DAgger (\figref{fig:ll_iact}), with both algorithms quickly learning a policy of quality equivalent to that of the expert. We list the full parameters of the algorithms in Tables 12 and 13. As in the previous section, \textbf{bold text} highlights sensitive hyperparameters.

\begin{table}[h]
\begin{center}
\begin{small}
\begin{sc}
\setlength{\tabcolsep}{2pt}
\begin{tabular}{lccccr}
\toprule
 Parameter & Value \\
\midrule
 Batch Size & 32 \\
 Gradient Steps $\pi$ Update & 3e3 \\
 \textbf{Gradient Steps $f$ Update} & \textbf{1e3} \\
 $f$ Gradient Penalty Target & 0 \\
 $f$ Gradient Penalty Weight & 5 \\
\bottomrule
\end{tabular}
\end{sc}
\end{small}
\end{center}
\caption{Learner hyperparameters for \texttt{DAeQuIL} on LunarLander-v2.}
\end{table}

\begin{table}[h]
\begin{center}
\begin{small}
\begin{sc}
\setlength{\tabcolsep}{2pt}
\begin{tabular}{lccccr}
\toprule
 Parameter & Value \\
\midrule
 Batch Size & 32 \\
 Gradient Steps $\pi$ Update & 1e4 \\
\bottomrule
\end{tabular}
\end{sc}
\end{small}
\end{center}
\caption{Learner hyperparameters for DAgger on LunarLander-v2.}
\end{table}

We next perform an experiment to show how careful curation of moments can allow \texttt{DAeQuIL} to significantly outperform DAgger at some tasks. Consider an operator trying to teach a drone to fly through a cluttered forest filled with trees. The operator has already trained a perception system that provides state information to the drone about whether a tree is infront of it. Because the operator is primarily concerned with safety, she only cares about making it through the forest, not the lateral location of the drone on the other side. 

She also tries to demonstrate a wide variety of evasive maneuvers as to hopefully teach the drone to generalize. We simulate such an operator and visualize the trajectories in \figref{fig:forest}, left.

Standard behavioral cloning with an $\ell_2$ loss would fail at this task because it would attempt to reproduce the conditional mean action, leading the drone to fly straight into the tree. Unfortunately, DAgger inherits this flaw, and is therefore prone to producing a policy that crashes into the first tree it sees, as shown in \figref{fig:forest}, center.

For \texttt{DAeQuIL}, the operator leverages her knowledge of the problem and passes in two important moments: the perception system's imminent crash indicator and the absolute difference between the current and proposed headings. Whenever the former is on, the latter is a large value under the expert's distribution as they are trying to avoid the tree. So, the learner figures out that it should swerve out of the way of the tree. This leads to policies learned via \texttt{DAeQuIL} to be able to progress much further into the forest, as seen in \figref{fig:forest}, right.

\begin{figure}[h]
    \centering
    \includegraphics[width=2\columnwidth/3]{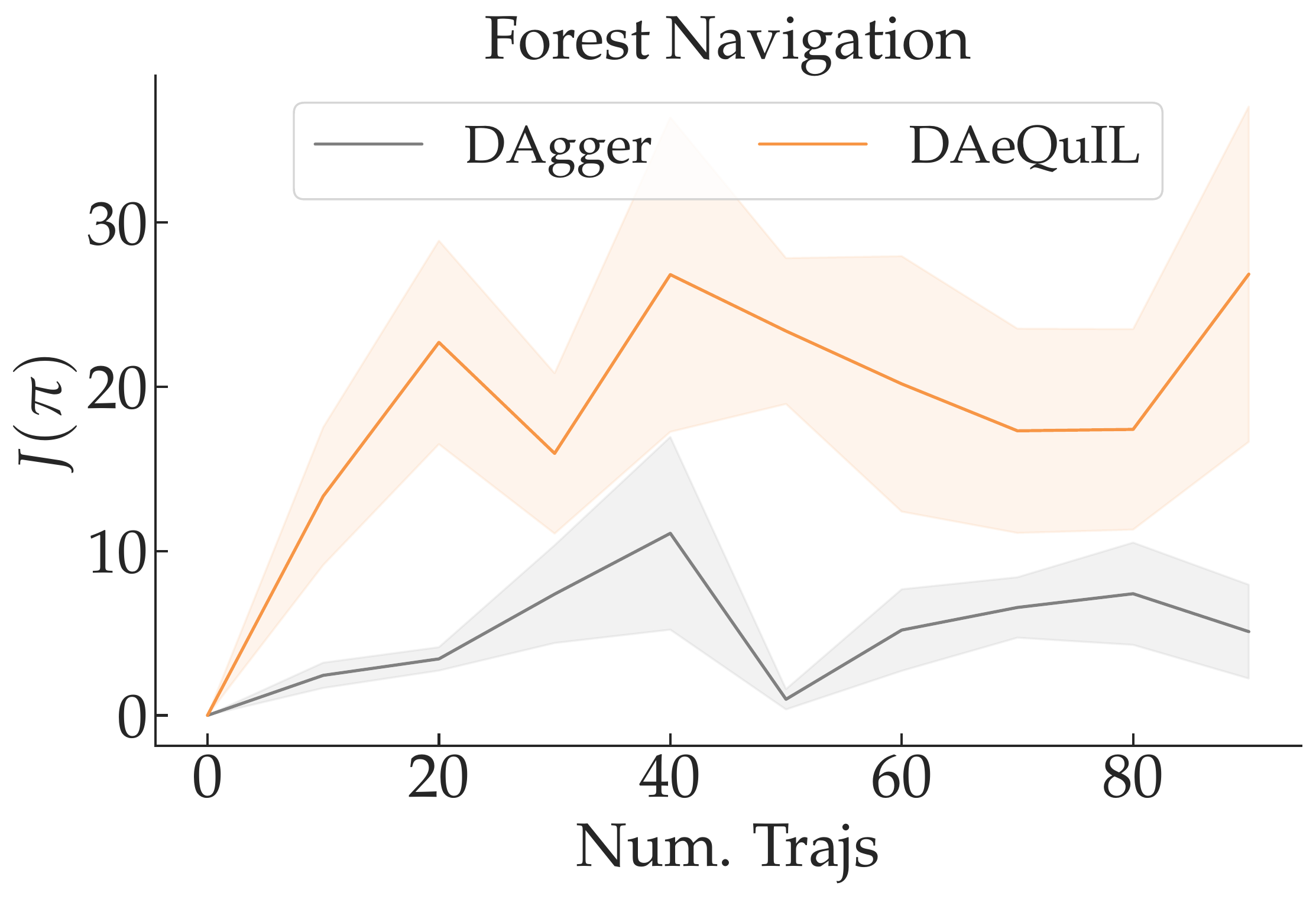}
    \caption{$J(\pi)$ is the longitudinal distance into the forest the learner is able to progress. All experiments are run on the forest layout shown in \figref{fig:forest} and standard errors are computed across 10 trials.}
    \label{fig:forest_iact}
\end{figure}

Using the final position of executed trajectories as the cumulative reward, we see the following learning curves with \texttt{DAeQuIL} clearly out-performing DAgger (\figref{fig:forest_iact}). 

We use the same function classes as the previous experiment but use a hidden size of $32$ for the discriminator of \texttt{DAeQuIL}. We list the full set of parameters in Tables 14 and 15.

\begin{table}[h]
\begin{center}
\begin{small}
\begin{sc}
\setlength{\tabcolsep}{2pt}
\begin{tabular}{lccccr}
\toprule
 Parameter & Value \\
\midrule
 Batch Size & 32 \\
 Gradient Steps $\pi$ Update & 2e3 \\
 \textbf{$\ell_{BC}$ Scale} & \textbf{5e-2} \\
 \textbf{Gradient Steps $f$ Update} & \textbf{1e3} \\
 $f$ Gradient Penalty Target & 0 \\
 $f$ Gradient Penalty Weight & 5 \\
\bottomrule
\end{tabular}
\end{sc}
\end{small}
\end{center}
\caption{Learner hyperparameters for \texttt{DAeQuIL} on Forest Navigation.}
\end{table}

\begin{table}[h]
\begin{center}
\begin{small}
\begin{sc}
\setlength{\tabcolsep}{2pt}
\begin{tabular}{lccccr}
\toprule
 Parameter & Value \\
\midrule
 Batch Size & 32 \\
 Gradient Steps $\pi$ Update & 5e3 \\
\bottomrule
\end{tabular}
\end{sc}
\end{small}
\end{center}
\caption{Learner hyperparameters for DAgger on Forest Navigation.}
\end{table}

\section{Additional Moment Types}
\label{sec:other_moments}
\subsection{A Fourth Moment Class: Mixed-Moment Value}
We could instead plug in $Q$-moments to the reward moment payoff function $U_1$. Let $\mathcal{F}_V$ and $\mathcal{F}_{V_E}$ refer to the classes of policy and expert value functions. As before, we assume both of these classes are closed under negation and include the true value and expert value functions. For notational convenience, we assume both classes contain functions with type signatures $\mathcal{S} \times \mathcal{A} \rightarrow \mathbb{R}$, with the second argument being ignored. Starting from the PDL, we can expand as follows:

\begin{flalign*}
  \begin{aligned}
&\quad  J(\pi_E) - J(\pi) \\[-4pt]
  &= \sum\nolimits_{t=1}^T \mathop{{}\mathbb{E}}_{\tau \sim \pi_E}[Q^{\pi}_t(s_t, a_t) - \mathop{{}\mathbb{E}}_{a \sim \pi(s_t)}[Q^{\pi}_t(s_t, a)]] \\[-4pt]
  &= \sum\nolimits_{t=1}^T \mathop{{}\mathbb{E}}_{\tau \sim \pi_E}[ Q^{\pi}_t(s_t, a_t) - \mathop{{}\mathbb{E}}_{a \sim \pi(s_t)}[Q^{\pi}_t(s_t, a)]] \\ &\hphantom{= \sum\nolimits_{t=1}^T} + \mathop{{}\mathbb{E}}_{\tau \sim \pi}[ Q^{\pi}_t(s_t, a_t) - Q^{\pi}_t(s_t, a_t)]\\[-4pt]
  &= \sum\nolimits_{t=1}^T \mathop{{}\mathbb{E}}_{\tau \sim \pi_E}[Q^{\pi}_t(s_t, a_t)]] - \mathop{{}\mathbb{E}}_{\tau \sim \pi}[Q^{\pi}_t(s_t, a_t)]] \\
  &\hphantom{= \sum\nolimits_{t=1}^T} + \mathop{{}\mathbb{E}}_{\substack{\tau \sim \pi \\ a \sim \pi(s_t)} }[Q^{\pi}_t(s_t, a)] - \mathop{{}\mathbb{E}}_{\substack{\tau \sim \pi_E \\ a \sim \pi(s_t)}}[Q^{\pi}_t(s_t, a)] \\
  & \leq \sup_{f \in \mathcal{F}_Q \cup \mathcal{F}_V} 2\sum\nolimits_{t=1}^T \mathop{{}\mathbb{E}}_{\tau \sim \pi}[f(s_t, a_t)] - \mathop{{}\mathbb{E}}_{\tau \sim \pi_E}[f(s_t, a_t)]
  \end{aligned}&
  \vspace{-4pt}
\end{flalign*}

The last step follows from the fact that $\sup_{a \in A} f(a) + \sup_{b \in B} f(b) \leq \sup_{c \in A \cup B} 2f(c)$. An analogous bound for $\mathcal{F}_{Q_E}$ and $\mathcal{F}_{V_E}$ can be proved by expanding the PDL in the reverse direction. We can use these expansions to provide bounds related to the reward-moment bound:

\begin{lemma}{\textbf{Mixed Moment Value Upper Bound:}} If $\mathcal{F}_{Q} / 2T$ and $\mathcal{F}_{V} / 2T$ spans $\mathcal{F}$ or $\mathcal{F}_{Q_E} / 2T$ and $\mathcal{F}_{V_E} / 2T$ do, then for all MDPs, $\pi_E$, and $\pi \leftarrow \Psi\{\epsilon\}(U_1)$, $J(\pi_E) - J(\pi) \leq O(\epsilon T^2)$.
\end{lemma}
\begin{pf} We start by expanding the imitation gap:
\vspace{-4pt}
\begin{flalign*}
  \begin{aligned}
    \quad & J(\pi_E) - J(\pi) \\[-4pt]
    & \leq \sup_{f \in \mathcal{F}_Q \cup \mathcal{F}_V} 2\sum\nolimits_{t=1}^T \mathop{{}\mathbb{E}}_{\tau \sim \pi}[f(s_t, a_t)] - \mathop{{}\mathbb{E}}_{\tau \sim \pi_E}[f(s_t, a_t)] \\[-4pt]
    & \leq \sup_{f \in \mathcal{F}} 2\mathop{{}\mathbb{E}}_{\tau \sim \pi}\sum\nolimits_{t=1}^T 2Tf(s_t, a_t) - \mathop{{}\mathbb{E}}_{\tau \sim \pi_E}\sum\nolimits_{t=1}^T 2Tf(s_t, a_t) \\[-4pt]
    & = 4T^2 \sup_{f \in \mathcal{F}} U_1(\pi, f) \leq 4 \epsilon T^2 
  \end{aligned}&
  \vspace{-4pt}
\end{flalign*}
The $T$ in the second to last line comes from the scaling down of either the $(\mathcal{F}_{Q}, \mathcal{F}_{V})$ or the $(\mathcal{F}_{Q_E}, \mathcal{F}_{V_E})$ pairs by $T$ to fit into the function class $\mathcal{F}$.
\end{pf}

\begin{lemma}{\textbf{Mixed Moment Value Lower Bound:}} There exists an MDP, $\pi_E$, and $\pi \leftarrow \Psi\{\epsilon\}(U_1)$ such that $J(\pi_E) - J(\pi) \geq \Omega(\epsilon T)$.
\label{bound:onq_low}
\end{lemma}
\begin{pf}
The proof of the \textit{reward lower bound} holds verbatim.
\end{pf}

These bounds show that solving this game, which might be more challenging than the reward-moment game, appears to offer no policy performance gains. However, in the imitation learning from observation alone setting, where one does not have access to action labels, reward-matching might be impossible, forcing one to use an approach similar to the above. This is because value functions are pure functions of state, not actions. \cite{sun2019provably} give an efficient algorithm for this setting.

\subsection{Combining Reward and Value Moments}
For both the \textit{off-$Q$} and \textit{on-$Q$} setups, one can leverage the standard expansion of a $Q$-function into a sum of rewards to derive a flexible family of algorithms that allow one to include knowledge of both reward and $Q$ moments. Explicitly, for the off-$Q$ case:
\begin{align}
    & J(\pi_E) - J(\pi) \nonumber \\[-4pt]
    &= \frac{1}{T}(\mathop{{}\mathbb{E}}_{\substack{\tau \sim \pi_E \\ a \sim \pi(s_{t})}}[\sum_{t=1}^T Q^{\pi}(s_t, a) - Q^{\pi}(s_t, a_t)])  \nonumber \\[-4pt]
    &= \frac{1}{T}(\mathop{{}\mathbb{E}}_{\substack{\tau \sim \pi_E \\ a \sim \pi(s_{t'})}}[\sum_{t=1}^T \sum_{t'=1}^{T'} r(s_{t'}, a) - r(s_{t'}, a_{t'}) \nonumber \\ &+ Q^{\pi}_{T'}(s_{T'}, a) - Q^{\pi}_{T'}(s_{T'}, a_{T'})])  \nonumber \\[-4pt]
    &\leq \max_{\substack{f \in \mathcal{F}_r \\ g \in \mathcal{F}_Q}} \frac{1}{T}(\mathop{{}\mathbb{E}}_{\substack{\tau \sim \pi_E \\ a \sim \pi(s_t)}}[\sum_{t=1}^T \sum_{t'=t}^{T'} f(s_{t'}, a) - f(s_{t'}, a_{t'})  \nonumber
    \\[-4pt] &+ g(s_{T'}, a) - g(s_{T'}, a_{T'})])
\end{align}

Passing such a payoff to our oracle with $\mathcal{F}$ spanned by $\mathcal{F}_r / 2 \times \mathcal{F}_Q / 2T$ would recover the off-$Q$ bounds.

This expansion begs the question of when it is useful. One answer is a standard bias/variance trade-off with different values of $T'$, as has been explored in TD-Gammon \cite{tesauro1995temporal}. We can provide an alternative answer by considering the limiting case -- when the $Q$ function is decomposed entirely into reward functions, the learner is required at timestep $t$ to match the sum of future reward moments. An efficient algorithm for such a problem can be derived as a natural extension of Policy Search by Dynamic Programming (PSDP) \cite{bagnell2003policy}, where, starting from $t = T-1$, the learner matches expert moments one timestep in the future, before moving one step backwards in time along the expert's trajectory. While this approach has the same performance characteristics as off-$Q$ algorithms, matching the class of reward moments might be simpler for some types of problems, like those with sparse rewards. However, it has the added complexity of producing a non-stationary policy.

We can perform an analogous expansion for the on-$Q$ case by utilizing the reverse direction of the PDL:
\begin{align}
    & J(\pi_E) - J(\pi) \nonumber \\[-4pt]
    &= \frac{1}{T}(\mathop{{}\mathbb{E}}_{\substack{\tau \sim \pi \\ a \sim \pi_E(s_{t})}}[\sum_{t=1}^T Q^{\pi_E}(s_t, a_t) - Q^{\pi_E}(s_t, a)])  \nonumber \\[-4pt]
    &= \frac{1}{T}(\mathop{{}\mathbb{E}}_{\substack{\tau \sim \pi \\ a \sim \pi_E(s_{t'})}}[\sum_{t=1}^T \sum_{t'=1}^{T'} r(s_{t'}, a_{t'}) - r(s_{t'}, a) \nonumber \\ &+ Q^{\pi_E}_{T'}(s_{T'}, a_{T'}) - Q^{\pi_E}_{T'}(s_{T'}, a)])  \nonumber \\[-4pt]
    &\leq \min_{\pi \in \Pi} \max_{\substack{f \in \mathcal{F}_r \\ g \in \mathcal{F}_{Q_E}}} \frac{1}{T}(\mathop{{}\mathbb{E}}_{\substack{\tau \sim \pi \\ a \sim \pi_E(s_{t'})}}[\sum_{t=1}^T \sum_{t'=t}^{T'} f(s_{t'}, a_{t'}) - f(s_{t'}, a)  \nonumber
    \\[-4pt] &+ g(s_{T'}, a_{T'}) - g(s_{T'}, a)])
\end{align}
Passing such a payoff to our oracle with $\mathcal{F}$ spanned by $\mathcal{F}_r / 2 \times \mathcal{F}_{Q_E} / 2T$ would recover the on-$Q$ bounds. A backwards-in-time dynamic-programming procedure is not possible for this expansion because of the need to sample trajectories from the policy at previous timesteps.

\end{document}